\documentclass{article}

\usepackage{arxiv}

\usepackage[utf8]{inputenc}
\usepackage[T1]{fontenc}
\usepackage{amssymb}
\usepackage{graphicx}
\usepackage{subcaption}
\usepackage{booktabs}
\usepackage{multirow}
\usepackage{longtable}
\usepackage{amsmath}
\usepackage{amssymb}
\usepackage{amsfonts}
\usepackage{bbm}
\usepackage{booktabs}
\usepackage{multirow}
\usepackage{graphicx}

\newcommand{\abms}[2]{#1$\pm$#2}
\usepackage{algorithm}
\usepackage{algpseudocode}
\usepackage{float}
\usepackage{placeins}
\usepackage{wrapfig}
\usepackage[numbers]{natbib}
\usepackage{url}
\usepackage{doi}
\usepackage{hyperref}
\usepackage{booktabs}
\usepackage{bibunits}
\usepackage{graphicx}
\usepackage{amssymb}
\usepackage[subtle,tracking=normal]{savetrees}
\def\ours{\texttt{LAViF}}
\newcommand{\ImgEnc}{\textsuperscript{\ensuremath{\dagger}}}
\newcommand{\VidEnc}{\textsuperscript{\ensuremath{\ddagger}}}

\title{Latent-Action-Guided Video-Language Feature Learning for Surgical Instrument–Tissue Interaction Recognition}

\author{
Jiajun Cheng$^{1}$,
Sainan Liu$^{3}$,
Subarna Tripathi$^{3}$,
Xiaofan Yu$^{2}$,
Shan Lin$^{1}$\\[4pt]
$^{1}$Arizona State University, Tempe, AZ, USA\\
$^{2}$University of California, Merced, CA, USA\\
$^{3}$Intel Corporation, USA\\[3pt]
\texttt{\{ccheng58, shan.lin.2\}@asu.edu}\\
\texttt{xiaofanyu@ucmerced.edu}\\
\texttt{\{subarna.tripathi, sainan.liu\}@intel.com}
}
\date{}
\begin{document}
\begin{bibunit}[plainnat]

\maketitle

\begin{abstract}
Recognizing instrument--tissue interactions is essential for
context-aware surgical AI. Vision-language models offer a natural way to
inject semantic structure into surgical representations by aligning video
features with textual action descriptions. However, pretrained encoders may
lack spatial coherence, while global semantic alignment does not ensure precise
spatial and temporal representations. By analyzing frame-to-frame feature
changes, we find that semantic alignment increases their dimensionality, but
larger increases do not necessarily improve recognition; encoders also differ
in how strongly dominant changes localize to interaction regions.
Motivated by these findings, we introduce \ours{}, which compresses
frame-to-frame changes into latent actions and predicts next-frame features
during end-to-end video--language alignment. Without additional spatial or
motion annotations, \ours{} improves the interaction grounding of leading
feature changes and temporal-direction sensitivity in our evaluated settings.
We further characterize how action capacity and prediction strength affect
recognition across encoders and triplet components. Using image encoders
without large-scale video pretraining, \ours{} achieves competitive recognition
with faster inference and smaller INT4 accuracy drops than V-JEPA2/2.1,
supporting its deployment potential. Project page: \url{https://marginlab.github.io/AI-for-healthcare/lavif/}.
 \vspace{-4mm}
\end{abstract}
\begin{center}
\begin{minipage}{0.9\columnwidth}
    \centering
    \includegraphics[width=0.9\linewidth]{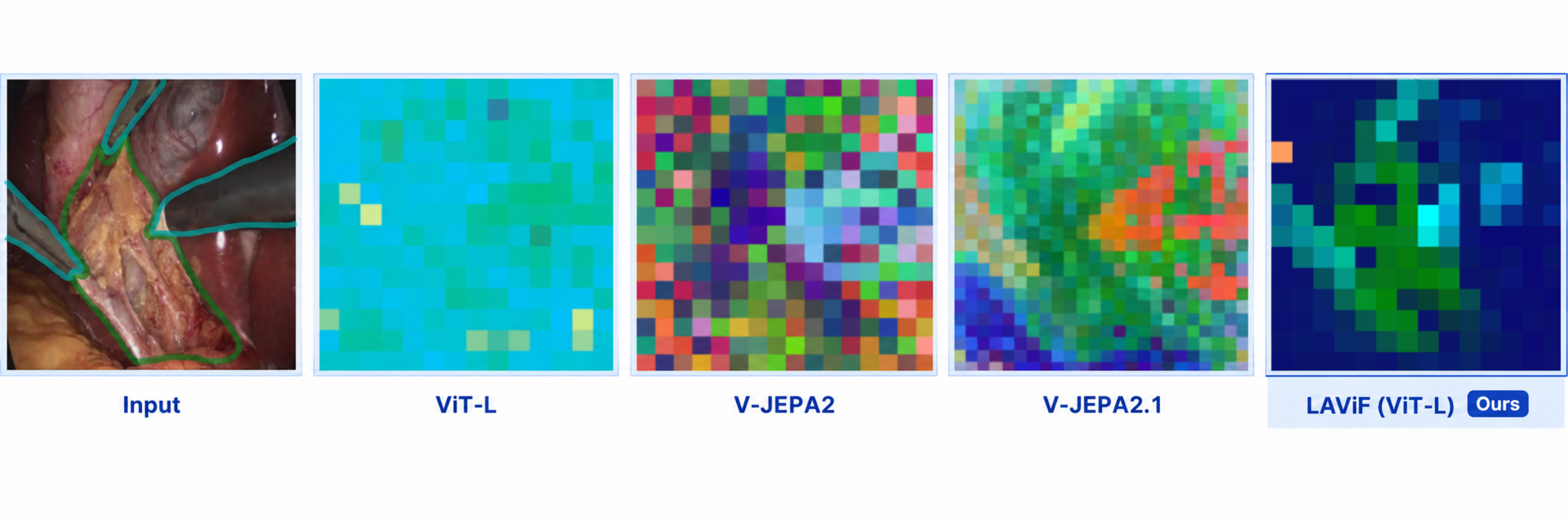}
    \vspace{-5mm}
    \captionsetup{type=figure,hypcap=false}
\caption{PCA-RGB maps encode the top three principal components
of patch features as RGB channels. Our method adapts a standard
ViT-L image backbone to produce spatially coherent maps
visually comparable to those of V-JEPA2/2.1, while showing clearer emphasis
on interacting instruments and target tissues and less
background detail in the displayed example.}
    \label{fig:intro}
\end{minipage}
\end{center}

\section{Introduction}

Artificial intelligence is increasingly being developed to make surgery more
context-aware, assistive, and autonomous
~\citep{yip2025robot,carstens2026artificial,cella2026future},
requiring robust scene understanding to interpret instruments, anatomy,
and their interactions in real time. Recognizing instrument--tissue
interactions represents what is being done, supporting context-aware
assistance and learning from surgical demonstrations.
This problem is commonly formulated as surgical action-triplet recognition,
which predicts
$\langle \mathrm{instrument}, \mathrm{verb}, \mathrm{target} \rangle$
triplet~\citep{nwoye2022data}.
Existing methods typically treat these components as categorical
prediction tasks~\citep{nwoye2022rdv,liu2024surgical,gui2024tail}.
While effective, closed-set supervision provides limited semantic
structure between related actions and can restrict generalization
beyond predefined categories.
v
Vision-language models (VLMs)
~\citep{radford2021learning,yuan2024hecvl,yuan2024procedure,yuan2025learning}
offer a complementary paradigm by aligning visual representations with
natural-language descriptions. Textual instrument-verb-target descriptions
provide a structured semantic space in which related triplets share
compositional structure~\citep{sharma2025fine,xi2023chain}. Language-aligned visual representations also offer a natural-language
interface for connecting perception with higher-level reasoning and planning
in large surgical AI systems~\citep{chen2023open,mao2026speak}.
More broadly, LaViLa~\citep{zhao2023lavila} and EgoVLP~\citep{lin2022egovlp}
demonstrate transfer of language-supervised video representations to
action recognition and retrieval. In surgery, descriptions can be
obtained from structured action labels or, at larger scale, narrated
video through automatic transcription and language processing
~\citep{yuan2025learning,yuan2024hecvl,yuan2024procedure}.
For downstream tasks like surgical action, pretrained
visual encoders can be kept frozen and their features matched
to text-derived representations
~\citep{rao2025zero,ikeido2026surgical}.
Alternatively, language supervision can guide adaptation of
the visual representations
~\citep{sharma2025fine,li2025recognize,che2026surglime,
walimbe2025adaptation}.
These approaches face two representation-level limitations.

\textbf{(1) Limitations inherited from frozen pretrained representations.}
Pretrained features can remain noisy or spatially imprecise
~\citep{khattak2023maple,menghini2023enhancing,wang2024sclip,el2024probing}.
Language-supervised pretraining provides strong semantics but does not
necessarily yield spatially coherent patch representations
~\citep{wang2025declip,lan2024clearclip}.
Self-supervised visual objectives
~\citep{wu2026unisurg,jamal2023surgmae,batic2023whether}
may recover stronger dense structure while retaining task-irrelevant
features~\citep{darcet2024vision,vanyan2023analyzing}.
Freezing the encoder preserves these limitations.

\textbf{(2) Global semantic alignment weakly constrains
interaction-relevant visual structure.}
End-to-end semantic alignment encourages global representations to
match target semantics but provides limited guidance on the local
evidence supporting that match. Encoders may exploit background context,
surgical phase, camera motion, or other correlated cues
~\citep{kumar2022fine,ding2022motion,choi2025goal,lan2024clearclip,zohra2026b},
without consistently organizing spatial and temporal feature changes
around the instrument--tissue interaction.

These limitations matter for surgical action recognition, which requires
localized interaction cues and temporal context
~\citep{nwoye2022rdv,lin2024instrument},
while complex and variable anatomy makes relevant structures difficult
to distinguish~\citep{madani2022guidance,honda2026capsule,cheng2026surgxbench}.
General-domain methods introduce spatial supervision through
detector-derived regions, region grouping, or segmentation guidance
~\citep{chen2024contrastive,pmlr-v162-zeng22c,sharma2025fine,zohra2026b,
choi2025goal}; surgical approaches similarly use instrument trajectories,
explicit instrument--tissue interaction guidance, or interaction
pseudo-labels~\citep{cheng2026trajpred,11084985,sharma2023surgical}.
However, automatically estimated spatial and motion cues can be unreliable
under occlusion, tissue deformation, and blurred anatomical boundaries.
We therefore aim to learn spatially grounded and temporally
direction-sensitive representations through semantic alignment,
without additional spatial or motion supervision.

The importance of localized action cues
~\citep{pei2025egovideo,wang2025actiondetail,long2025bdcclip}
and temporal direction~\citep{bagad2025chirality}
motivates analyzing \emph{transition structure}: how frame-to-frame
feature changes are distributed across feature-space directions
and where those directions are spatially grounded.
Let $s_t\in\mathbb{R}^{d_s}$ denote the visual feature vector
(state) of frame $t$. We define the transition
$\delta_t=s_{t+1}-s_t$ and analyze the spectrum of its covariance
$\Sigma_\delta=\operatorname{Cov}(\delta)$ under semantic alignment training.
Across the evaluated encoders, semantic alignment expands the effective
transition space, but greater expansion does not consistently imply
better recognition or stronger temporal-direction sensitivity.
Instead, strong recognition can coexist with a compact transition
space whose dominant changes are grounded in instrument--tissue interactions.
This motivates investigating whether such transition structure
can be encouraged during semantic alignment.

We therefore investigate shaping transition structure through joint
predictive and semantic learning.
Prior analysis shows that next-state prediction through compact latent
actions can preserve dominant transition directions in fixed
representations~\citep{zhang2026latent}.
Extending this setting to a jointly trained encoder, \ours{} uses a
\emph{Causal Transition IDM} to compress current and preceding
transitions into action tokens.
These tokens condition next-state prediction through a forward dynamics
model (FDM) and join frame states for video--language alignment,
supporting both dynamics and interaction semantics.
A patch-level SIG regularizer further stabilizes local features during joint training.
Related inverse/forward dynamics and latent-action work is discussed
in Section \ref{sec: related work}.

Our contributions are threefold:
\textbf{(1) Empirical finding.}
We show that semantic alignment reshapes visual transition structure:
strong recognition can coincide with compact spectra and
interaction-grounded leading directions, whereas greater dimensional
expansion alone does not imply better recognition
(Section~\ref{sec:transition_analysis}).
\textbf{(2) Finding-driven method.}
We introduce \ours{} to investigate and explicitly shape this structure
through a Causal Transition IDM, a compact predictive action bottleneck,
video--language alignment, and patch-level SIG regularization (SIGReg)
(Section~\ref{sec:components}).
\textbf{(3) Empirical validation.}
Across encoders and datasets, we show that \ours{} can improve interaction
grounding, temporal-direction sensitivity, and recognition without
additional spatial or motion supervision, and characterize the effects
of bottleneck capacity and prediction strength
(Section~\ref{sec:exp}).
We also demonstrate faster inference and smaller performance drops under
INT4 quantization than the evaluated pretrained video encoders.

\section{Transition Structure under Semantic Alignment Training}
\label{sec:transition_analysis}

We first characterize how semantic alignment training reshapes frame-to-frame visual
transitions, and organize our analysis around the following empirical observations.

\paragraph{Setup.}
We use CholecT50~\citep{nwoye2022data}, a laparoscopic cholecystectomy
action-triplet benchmark. For each annotated frame $x_t$, we form an
eight-frame clip $\{x_{t-7},\ldots,x_t\}$ from the 1\,fps sequence and
predict its instrument--verb--target labels at $x_t$. All encoders use the
same temporal context, video--text objective, and EmbeddingGemma text
encoder~\citep{vera2025embeddinggemma}. Image encoders
(ViT-L~\citep{dosovitskiy2021image}, DINOv2-L~\citep{oquab2023dinov2},
SurgeNet-L~\citep{JASPERS2026103873}) process frames independently followed
by a CLIP4Clip-style temporal Transformer~\citep{luo2021clip4clip}, whereas
native video encoders (TimeSformer, V-JEPA2~\citep{assran2025vjepa2}, and
V-JEPA2.1~\citep{mur2026v}) process clips jointly. For patch features $\phi_p^t$, we define
$s_t=\frac{1}{N}\sum_{p=1}^{N}\phi_p^t$ and
$\delta_t=s_{t+1}-s_t$. We use \emph{transition structure} to denote
\textbf{(1)} how transition variance is distributed across feature-space
directions and \textbf{(2)} where those directional changes occur spatially.
We estimate $\Sigma_\delta=\operatorname{Cov}(\delta)$ from $20{,}000$
consecutive-frame pairs and summarize its spectrum by $d_{90}$, the minimum
number of leading directions explaining $90\%$ of transition variance.
Spatial grounding is measured by projecting patch transitions
$\phi_p^{t+1}-\phi_p^t$ onto covariance-eigenvector bands, while
temporal-direction sensitivity is measured by the relative recognition drop
under clip reversal.

% Preamble: \usepackage{booktabs,caption}

% Requires \usepackage{booktabs,caption}
% Put \raggedbottom immediately after \begin{document}.

\begin{figure*}[t]
\centering

\begin{minipage}[t]{0.66\textwidth}
\vspace{0pt}
\centering
{\small\bfseries (a) Transition dimensionality, recognition, and reversal\par}
\vspace{4pt}

\scriptsize
\renewcommand{\arraystretch}{0.90}
\setlength{\tabcolsep}{2.5pt}
\setlength{\aboverulesep}{1pt}
\setlength{\belowrulesep}{1pt}

\resizebox{\linewidth}{!}{%
\begin{tabular}{
    @{}l
    c@{\hspace{5pt}}
    cccc@{\hspace{5pt}}
    cccc@{}
}
\toprule
& $d_{90}$
& \multicolumn{4}{c}{\textbf{mAP (\%)}}
& \multicolumn{4}{c}{\textbf{Reversal drop (\%)}} \\
\cmidrule(lr){3-6}
\cmidrule(lr){7-10}
Encoder & Before $\rightarrow$ after
& IVT & I & V & T
& $\Delta_{\mathrm{IVT}}$ & $\Delta_{\mathrm{I}}$
& $\Delta_{\mathrm{V}}$ & $\Delta_{\mathrm{T}}$ \\
\midrule

\multicolumn{10}{@{}l}{\textit{Image encoders}} \\
DINOv2-L
& $278\rightarrow391$
& 33.52 & 94.26 & 67.26 & 48.09
& 19.6 & 13.1 & 13.8 & 16.1 \\
ViT-L
& $406\rightarrow537$
& 31.29 & 91.66 & 66.41 & 45.56
& 21.8 & 13.9 & 15.1 & 13.0 \\
SurgeNet-L
& $261\rightarrow404$
& 33.09 & 94.29 & 69.62 & 48.58
& 28.2 & 15.5 & \textbf{19.0} & 14.7 \\
\midrule

\multicolumn{10}{@{}l}{\textit{Video encoders}} \\
TimeSformer
& $333\rightarrow431$
& 27.32 & 88.67 & 64.69 & 44.43
& 5.1 & 3.8 & 3.4 & 4.2 \\
V-JEPA2
& $80\rightarrow231$
& 29.68 & 92.16 & 68.55 & 47.97
& 28.5 & 14.0 & 16.5 & 17.2 \\
V-JEPA2.1
& $\mathbf{45\rightarrow203}$
& \textbf{34.82} & \textbf{93.99}
& \textbf{69.76} & \textbf{50.23}
& \textbf{30.0} & \textbf{17.7} & 17.0 & 14.4 \\
\bottomrule
\end{tabular}%
}
\end{minipage}\hfill
\begin{minipage}[t]{0.31\textwidth}
\vspace{0pt}
\centering
{\small\bfseries (b) Spatial grounding of transitions\par}
\vspace{4pt}
\includegraphics[width=\linewidth]{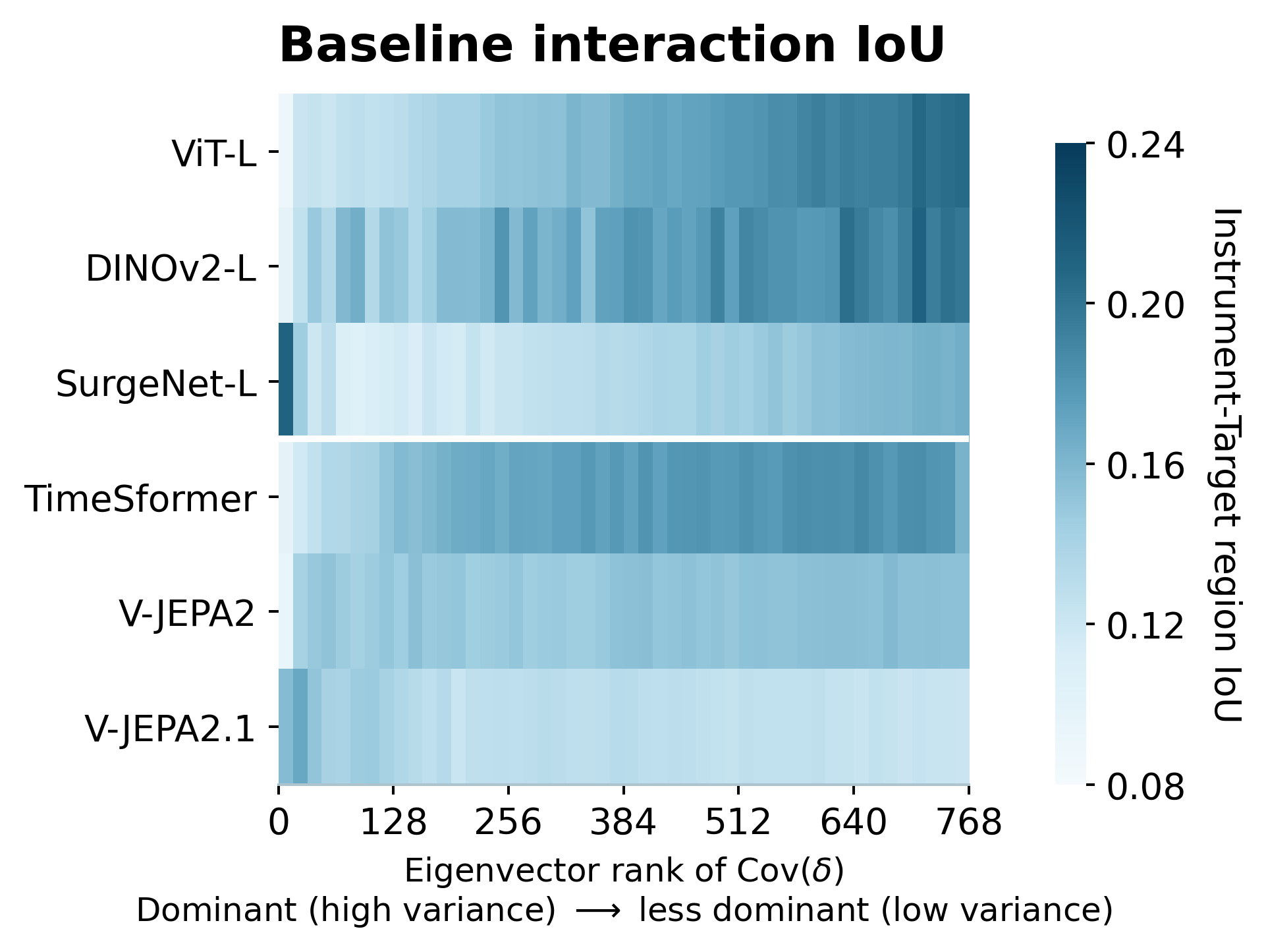}
\end{minipage}

\caption{
\textbf{Transition structure on CholecT50.}
\textbf{(a)} $d_{90}$ before and after semantic alignment,
 mAP after training, and relative mAP drop under clip reversal.
\textbf{(b)} Per-band IoU with the instrument--target region;
higher-variance transition directions appear on the left.
}
\label{fig:transition_overview}
\end{figure*}

\textbf{Observation 1: A larger transition space does not consistently
imply better recognition or stronger temporal-direction sensitivity.}
Semantic alignment increases $d_{90}$ for every evaluated encoder
(Figure~\ref{fig:transition_overview}(a)).
After training, ViT-L and TimeSformer have larger $d_{90}$
($537$ and $431$) than V-JEPA2.1 ($203$), yet lower
$\mathrm{mAP}_{IVT}$ ($31.29$ and $27.32$ versus $34.82$)
and smaller reversal drops ($21.8\%$ and $5.1\%$ versus
$30.0\%$). Thus, transition dimensionality alone does not explain recognition or temporal-direction sensitivity.

 %We therefore hypothesize that well-structured pretrained
%representations can encode fine-grained interaction changes without recruiting
%a large number of additional transition directions.

\textbf{Observation 2: Interaction grounding of transition structure varies
across encoders and transition ranks.}
We rank the eigenvectors of $\Sigma_\delta$ by decreasing variance and group
them into bands of $16$. Each patch is scored by the fraction of its
frame-to-frame feature-change energy captured by a band; the highest-scoring
$25\%$ are foreground and the remainder background, forming a binary
saliency map. We compute its IoU with the instrument--target region formed
by combining SASVi-estimated segmentation masks from both
frames~\citep{sivakumar2025sasvi}, averaging over $5{,}000$ evaluation
frames. Projection and mask-construction details are provided in
Appendix A.3.
Figure~\ref{fig:transition_overview} (a, b) shows that V-JEPA2.1,
with the highest $\mathrm{mAP}_{IVT}$ ($34.82$), has higher IoU in its
leading bands than its tail; TimeSformer shows the opposite trend with
lower recognition ($27.32$). DINOv2-L exceeds ViT-L in several leading
bands, while SurgeNet-L exhibits a pronounced first-band peak; both
outperform ViT-L in recognition. Notably, these self-distillation-based
encoders (DINOv2, SurgeNet, V-JEPA2.1) exhibit stronger localization in their dominant directions of transition structure.
Together, these observations suggest that a larger effective transition space does not necessarily imply that more transition directions capture action-relevant information, as some may instead reflect task-irrelevant variation. This motivates the hypothesis that transition directions of different visual encoders may differ in their task relevance, and that stronger localization of dominant directions to interaction regions may be associated with better surgical action recognition.

These observations on the transition structure raise a question: \emph{Can transition structure be
explicitly shaped during semantic alignment so that dominant visual changes
better reflect instrument--tissue interactions while mitigating irrelevant variation?}
Recent work~\citep{zhang2026latent} shows that compressing transitions into a low-dimensional latent action for next-state prediction can preserve dominant transition directions when visual features are fixed.
We investigate whether jointly training the encoder with this training objective can
reshape the transition structure and improve where they are grounded, without
additional spatial or motion annotations.

\section{Latent-Action-Guided Design}
\label{sec:components}

\begin{figure}[t]
    \centering
    \includegraphics[width=\linewidth]{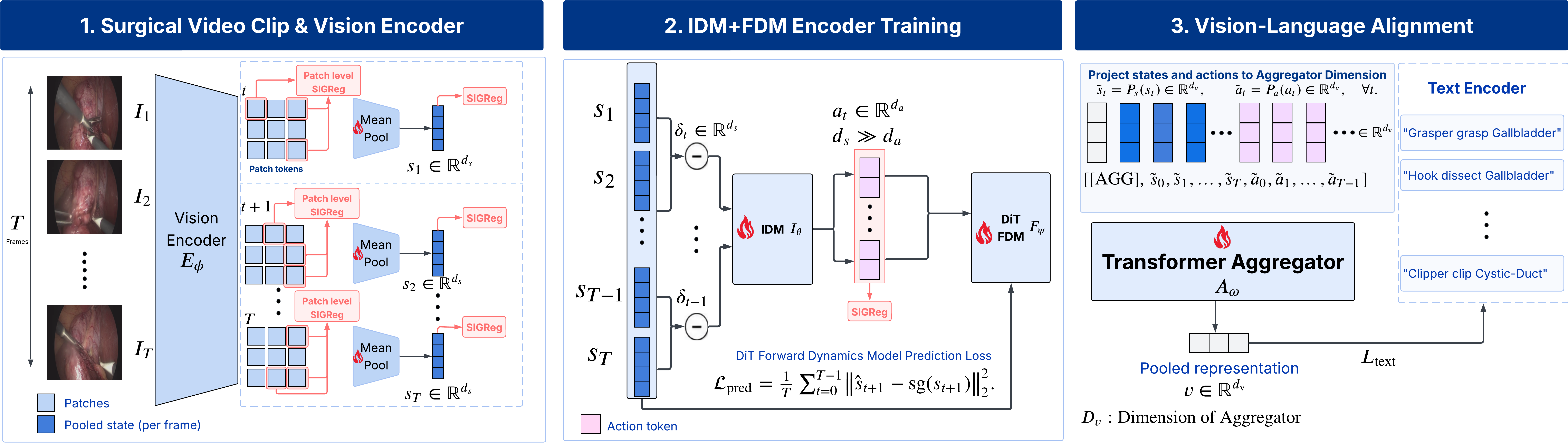}
    \caption{
\textbf{Overview of LAViF.}
From left to right: \textbf{(1)} the surgical video is encoded into frame-level visual states;
\textbf{(2)} the causal IDM maps state transitions to latent actions, which condition FDM next-state prediction;
and \textbf{(3)} state--action tokens are aggregated and aligned with text embeddings for surgical interaction recognition.
SIGReg regularizes state, action, and patch features.
}
    \label{fig:pipeline}
\end{figure}

To investigate whether transition structure can be explicitly shaped during semantic alignment, and whether doing so benefits surgical action recognition, we propose \ours{}. Building on latent-action modeling for predictive representation learning and control~\citep{zhang2026latent,ye2025lapa,schmidt2024lapo,cui2024dynamo}, \ours{} introduces compact latent actions that jointly support next-state prediction and semantic interaction recognition. The framework consists of a vision encoder $E_\phi$, an inverse dynamics model (IDM) $I_\theta$, a Forward Dynamics Model (FDM) $F_\psi$, a clip aggregator $A_\omega$, and a text encoder $G_\chi$ (Figure~\ref{fig:pipeline}).

\paragraph{Vision Encoder.}
For each frame, $E_\phi$ produces patch tokens $\mathbf{p}_t$ and a pooled visual state
$s_t=\operatorname{pool}(\mathbf{p}_t)\in\mathbb{R}^{d_s}$.
All visual representations are jointly adapted during training.

\paragraph{Causal Transition Inverse Dynamics Model.}
We represent each state transition by
$\delta_t=s_{t+1}-s_t$, following displacement-based action decoding~\citep{zhang2026delta}.
The IDM processes the transition sequence with a causally masked Transformer and outputs
\begin{equation}
a_t=I_\theta(\delta_{\leq t})\in\mathbb{R}^{d_a},
\qquad d_a\ll d_s.
\end{equation}
The causal mask incorporates transition history while preserving temporal asymmetry, and the low-dimensional $a_t$ forms a compact bottleneck for transition-specific information.

\paragraph{Forward Dynamics Model.}
The FDM predicts the next visual state from the current state history conditioned on the latent action,
$\hat{s}_{t+1}=F_\psi(s_{\leq t},a_t)$.
We train it with a stop-gradient prediction objective,
\begin{equation}
\mathcal{L}_{\mathrm{pred}}
=\frac{1}{T}\sum_{t=0}^{T-1}
\left\|\hat{s}_{t+1}-\operatorname{sg}(s_{t+1})\right\|_2^2.
\end{equation}
Because transition-specific information is routed through the compact $a_t$, this objective encourages the latent action to retain predictive state changes. The FDM uses a causal DiT-style architecture with adaLN-zero conditioning~\citep{peebles2023scalable,maes2026lewm}.

\paragraph{SIG Regularization.}
We apply SIGReg~\citep{maes2026lewm,balestriero2025lejepa} to state,
action, and patch representations. Specifically,
\[
\mathcal{L}_{\mathrm{SIG}}
=
\operatorname{SIGReg}(\{s_t\})
+
\operatorname{SIGReg}(\{a_t\}),
\qquad
\mathcal{L}^{\mathrm{patch}}_{\mathrm{SIG}}
=
\operatorname{SIGReg}
\bigl(\{p_{t,k}:k\in\mathcal{K}_t\}\bigr),
\quad |\mathcal{K}_t|=32,
\]
where $\mathcal{K}_t$ contains 32 patch tokens sampled uniformly from frame $t$,
resampled independently for each clip and frame at every training step. State
regularization prevents representation collapse, action regularization
discourages the FDM from ignoring the latent action, and patch-level SIGReg
preserves local feature diversity during adaptation. The SIGReg statistic and
projection settings are given in Appendix~A.4.

% \paragraph{Aggregation and Semantic Alignment.}
% The aggregator projects state and action tokens into a shared space and processes
% $[\texttt{[AGG]},s_0,\dots,s_T,a_0,\dots,a_{T-1}]$
% with a bidirectional Transformer. The resulting clip embedding therefore combines
% states (\emph{what is present}) and latent actions (\emph{what is changing}).
% Importantly, the IDM receives semantic supervision through the recognition pathway while simultaneously receiving predictive supervision through the FDM, allowing the shared action bottleneck to be optimized for both objectives.

\paragraph{Aggregation and Semantic Alignment.}
The aggregator $A_\omega$ linearly projects state and action tokens, prepends a learnable $\texttt{[AGG]}$ token, and processes the full sequence with a bidirectional Transformer:
$
v=A_\omega\big([\texttt{[AGG]},s_0,\dots,s_T,a_0,\dots,a_{T-1}]\big).
$
Combining states (\emph{what is present}) and latent actions (\emph{what is changing}) allows the IDM to receive semantic supervision through the recognition pathway together with predictive supervision from the FDM, while jointly adapting the visual encoder.
EmbeddingGemma~\citep{vera2025embeddinggemma} encodes complete instrument--verb--target descriptions and their individual components. We additionally supervise instrument, verb, and target semantics, following prior work that uses component-level supervision for fine-grained action and video--language learning~\citep{wray2019fine,momeni2023verbs,vo2024henasy,sharma2025fine}. Component labels are obtained directly from the fixed triplet-to-component mapping.
Temperature-scaled cosine similarity between clip and text embeddings produces triplet and component logits, optimized with binary cross-entropy:
\[
\mathcal{L}_{\mathrm{text}}
=\lambda_{\mathrm{IVT}}\mathcal{L}_{\mathrm{IVT}}
+\sum_{q\in\{\mathrm{I},\mathrm{V},\mathrm{T}\}}\lambda_q\mathcal{L}_q .
\]
The full objective is
\[
\mathcal{L}
=\mathcal{L}_{\mathrm{text}}
+\lambda_{\mathrm{pred}}\mathcal{L}_{\mathrm{pred}}
+\lambda_s\mathcal{L}_{\mathrm{SIG}}
+\lambda_p\mathcal{L}_{\mathrm{SIG}}^{\mathrm{patch}}.
\]

The FDM is used only during training; at inference, $E_\phi$, $I_\theta$, and $A_\omega$ produce the clip representation matched against instrument--verb--target text embeddings. We use $\lambda_{\mathrm{pred}}=1$ for the full model, while $\lambda_{\mathrm{pred}}=0$ retains the IDM and state--action recognition pathway. Unless explicitly stated otherwise, all controlled baselines and ablations are trained with the same BCE-based semantic alignment objective.
Additional architectural and optimization details are provided in Appendix A.4.

\section{Experiments}
\label{sec:exp}
\begin{table*}[t]
\centering
\scriptsize
\setlength{\tabcolsep}{3.4pt}
\renewcommand{\arraystretch}{1.08}

\caption{
Main comparison on CholecT50 and ProstaTD with $d_a=128$.
\ImgEnc{} denotes frame-wise image encoders and
\VidEnc{} denotes native video encoders.
$\triangle$ marks methods using a classification head without
video--language alignment.
$\diamond$ marks methods evaluated with different metrics and
therefore requiring cautious comparison.
All remaining baselines and \ours{} variants are trained under
our controlled training protocol.
Best and second-best results are shown in \textbf{bold} and
\underline{underline}, respectively.
}
\label{tab:main_results}

\resizebox{\textwidth}{!}{%
\begin{tabular}{lcccccccc}
\toprule

\textbf{Method}
& \multicolumn{4}{c}{\textbf{CholecT50 (RDV)}}
& \multicolumn{4}{c}{\textbf{ProstaTD (5-fold)}} \\
\cmidrule(lr){2-5}
\cmidrule(lr){6-9}

& $\mathbf{mAP_{IVT}}$
& $\mathbf{mAP_I}$
& $\mathbf{mAP_V}$
& $\mathbf{mAP_T}$
& $\mathbf{mAP_{IVT}}$
& $\mathbf{mAP_I}$
& $\mathbf{mAP_V}$
& $\mathbf{mAP_T}$ \\
\midrule

% ==========================================================
% Published task-specific methods
% ==========================================================
\multicolumn{9}{l}{
\textit{\textbf{Published task-specific methods}}} \\
\addlinespace[1pt]

Rendezvous~\citep{nwoye2022rdv}\ImgEnc $^\triangle$
& 29.90 & 92.20 & 60.07 & 38.30
& -- & -- & -- & -- \\

Chain-of-Look~\citep{xi2023chain}\ImgEnc $^\triangle$
& \textbf{38.00} & 94.10 & 62.50 & 41.90
& -- & -- & -- & -- \\

MML-SurgAdapt~\citep{walimbe2025adaptation}\ImgEnc
& 30.30 & 87.30 & 61.10 & 43.20
& -- & -- & -- & -- \\

TDNet~\citep{prostatd}\ImgEnc$^\diamond$$^\triangle$
& -- & -- & -- & --
& 36.10 $\pm$ 3.40
& 89.90 $\pm$ 1.30
& 61.70 $\pm$ 2.90
& 55.70 $\pm$ 2.40 \\

\midrule

% ==========================================================
% Controlled baselines
% ==========================================================
\multicolumn{9}{l}{
\textit{\textbf{Controlled baselines}}} \\
\addlinespace[1pt]

\multicolumn{9}{l}{\hspace{0.8em}\textit{General-domain pretraining}} \\

CLIP ViT-L/14\ImgEnc
& 26.17 & 86.26 & 62.54 & 42.93
& 26.80 $\pm$ 0.96
& 90.97 $\pm$ 0.76
& 65.00 $\pm$ 3.28
& 53.51 $\pm$ 2.71 \\

DINOv2-L\ImgEnc
& 27.41 & 90.08 & 65.28 & 45.44
& 36.01 $\pm$ 1.95
& 93.13 $\pm$ 0.72
& 70.75 $\pm$ 2.78
& 63.84 $\pm$ 2.32 \\

V-JEPA2\VidEnc
& 29.68 & 92.16 & 68.55 & 47.97
& 32.66 $\pm$ 1.85
& 91.30 $\pm$ 1.27
& 71.89 $\pm$ 3.29
& 59.95 $\pm$ 1.78 \\

V-JEPA2.1\VidEnc
& \underline{34.82}
& 93.99
& \underline{69.76}
& \underline{50.23}
& 35.87 $\pm$ 2.50
& 94.36 $\pm$ 1.80
& \textbf{74.35} $\pm$ 2.47
& 62.64 $\pm$ 2.47 \\

TimeSformer\VidEnc
& 27.32 & 88.67 & 64.69 & 44.43
& 29.43 $\pm$ 2.11
& 86.19 $\pm$ 1.71
& 64.69 $\pm$ 3.19
& 56.14 $\pm$ 3.25 \\

\addlinespace[2pt]
\multicolumn{9}{l}{\hspace{0.8em}\textit{Surgical-domain pretraining}} \\

EndoViT\ImgEnc
& 24.71 & 85.19 & 61.41 & 41.23
& 26.94 $\pm$ 0.89
& 88.35 $\pm$ 0.82
& 62.02 $\pm$ 3.97
& 53.63 $\pm$ 2.44 \\

LEMON\ImgEnc
& 28.71 & 87.34 & 65.04 & 44.70
& 32.26 $\pm$ 1.22
& 91.72 $\pm$ 1.33
& 66.51 $\pm$ 3.29
& 59.71 $\pm$ 1.41 \\

SurgVLP\ImgEnc
& 20.28 & 73.96 & 52.33 & 37.76
& 20.64 $\pm$ 0.30
& 78.07 $\pm$ 1.73
& 52.45 $\pm$ 2.29
& 45.88 $\pm$ 2.36 \\

HecVLP\ImgEnc
& 21.41 & 77.59 & 57.23 & 39.29
& 22.42 $\pm$ 0.42
& 81.16 $\pm$ 2.68
& 55.89 $\pm$ 2.70
& 48.76 $\pm$ 2.65 \\

PeskaVLP\ImgEnc
& 22.02 & 79.20 & 57.22 & 39.26
& 23.09 $\pm$ 0.64
& 82.45 $\pm$ 2.01
& 56.99 $\pm$ 3.44
& 50.75 $\pm$ 3.34 \\

\midrule

% ==========================================================
% Ours
% ==========================================================
\addlinespace[1pt]

\ours{} -- ViT-L
& 32.15 & 93.24 & 68.29 & 47.05
& 32.82 $\pm$ 1.29
& 92.82 $\pm$ 1.00
& 68.28 $\pm$ 2.80
& 60.63 $\pm$ 1.73 \\

\ours{} -- DINOv2-L
& 33.13
& \underline{94.60}
& 68.91
& 48.81
& \underline{38.14} $\pm$ 2.61
& \underline{96.03} $\pm$ 2.19
& \underline{73.01} $\pm$ 2.98
& \underline{65.21} $\pm$ 2.92 \\

\ours{} -- SurgeNet-L
& 32.78
& \textbf{95.02}
& \textbf{71.64}
& \textbf{50.89}
& \textbf{38.85} $\pm$ 2.07
& \textbf{96.74} $\pm$ 1.24
& 72.82 $\pm$ 1.32
& \textbf{65.73} $\pm$ 2.38 \\

\bottomrule
\end{tabular}
}
\end{table*}
We first describe the datasets, baselines, ablations, and metrics.
Section~\ref{sec:recognition} evaluates recognition, component ablations,
temporal-direction sensitivity, and deployment efficiency. The default full
model uses $d_a=128$ and $\lambda_{\mathrm{pred}}=1$.
Section~\ref{sec:enc-behavior} varies $d_a$ and $\lambda_{\mathrm{pred}}$
to study transition structure and recognition, using the transition-band IoU
from Section~\ref{sec:transition_analysis} and PCA-RGB maps to examine spatial
grounding and local feature organization.

\paragraph{Datasets.}
We evaluate on two surgical action-triplet benchmarks.
\textbf{CholecT50}~\citep{nwoye2022data} contains 50 laparoscopic
cholecystectomy videos (90,489 frames at 1\,fps) and 100 triplet classes
over 6 instruments, 10 verbs, and 15 targets; we use the standard Rendezvous
(RDV) split.
\textbf{ProstaTD}~\citep{prostatd} contains 21 robot-assisted radical
prostatectomy videos ($\sim$72K frames) and 89 triplet classes over
7 instruments, 10 verbs, and 10 targets; we use its five-fold protocol.
Together they span general surgery and urology, including hepatobiliary
dissection and crowded pelvic scenes with difficult tissue-layer identification
~\citep{madani2022guidance,honda2026capsule,prostatd}.
For each labeled frame $x_t$, we form
$\mathbf{X}_t=\{x_{t-7},\ldots,x_t\}$ and assign it the label at $x_t$.
Image baselines use only $x_t$, native video encoders use the full clip, and
\ours{} constructs an eight-frame state--action representation from pretrained
image backbones.

\paragraph{Baselines.}
We compare general-domain encoders
CLIP~\citep{radford2021learning}, DINOv2~\citep{oquab2023dinov2},
V-JEPA2/2.1~\citep{assran2025vjepa2,mur2026v}, and
TimeSformer~\citep{bertasius2021space};
surgical-domain encoders EndoViT~\citep{batic2023whether}, LEMON~\citep{che2026lemon},
SurgVLP~\citep{yuan2025learning}, HecVLP~\citep{yuan2024hecvl}, and
PeskaVLP~\citep{yuan2024procedure}; and task-specific methods
Rendezvous~\citep{nwoye2022rdv},
Chain-of-Look~\citep{xi2023chain}, and
MML-SurgAdapt~\citep{walimbe2025adaptation}.
We include V-JEPA2 as a strong pretrained video encoder and V-JEPA2.1 as its
enhanced variant, which further incorporates self-distillation to promote
more spatially coherent, object-level representations.
The task-specific methods use their original frameworks and serve as reference
comparisons; Chain-of-Look also uses task-specific prior information.
We instantiate \ours{} with ViT-L/16, DINOv2-L/16, and
SurgeNet-L/14~\citep{JASPERS2026103873}, representing vanilla, general-domain
self-supervised, and surgical-domain self-supervised pretraining. These models
and controlled baselines are fully trainable with the same EmbeddingGemma text
encoder~\citep{vera2025embeddinggemma} and BCE-based semantic alignment objective
from Section~\ref{sec:components}. Baseline and training details are in
Appendix~A.2.

\paragraph{Ablations and metrics.}
Within each backbone and dataset, all variants share data splits, learning rates,
training schedules, epoch budgets, and semantic objectives. We compare the
no-IDM+FDM baseline, IDM only, IDM+FDM without patch-level SIGReg, and the full
model. IDM only disables both forward prediction and patch-level SIGReg;
the no-patch-SIGReg variant isolates the regularizer. Note that without IDM+FDM, image encoders use frame-wise features followed by a
CLIP4Clip-style temporal Transformer~\citep{luo2021clip4clip}, as in
Section~\ref{sec:transition_analysis}.
We report $\mathrm{mAP}_{IVT}$ and component-level
$\mathrm{mAP}_{I}$, $\mathrm{mAP}_{V}$, and $\mathrm{mAP}_{T}$
using \emph{ivtmetrics}~\citep{nwoye2022data}; ProstaTD reports mean
$\pm$ standard deviation over five folds. Deployment evaluation measures
per-clip latency, throughput, peak GPU memory, and $\mathrm{mAP}_{IVT}$
changes from FP32 under INT8/INT4 post-training weight-only quantization
without fine-tuning.

% ============================================================
% MAIN COMPARISON
% ============================================================

% ============================================================
% MAIN PERFORMANCE
% ============================================================
\begin{table*}[t]
\centering
\caption{
Component ablations at $d_a=128$.
Values are recognition mAP (\%); ProstaTD reports
mean$_{\pm\mathrm{std}}$ over five folds.
}
\label{tab:combined_ablation}

\scriptsize
\renewcommand{\arraystretch}{0.92}
\setlength{\tabcolsep}{3pt}
\setlength{\aboverulesep}{1pt}
\setlength{\belowrulesep}{1pt}

\resizebox{\textwidth}{!}{%
\begin{tabular}{@{}ll cccc cccc@{}}
\toprule
& & \multicolumn{4}{c}{\textbf{CholecT50}}
  & \multicolumn{4}{c}{\textbf{ProstaTD}} \\
\cmidrule(lr){3-6}\cmidrule(lr){7-10}
Backbone & Variant
& IVT & I & V & T
& IVT & I & V & T \\
\midrule

\multirow{4}{*}{ViT-L}
& Full
& 32.15 & 93.24 & 68.29 & 47.05
& \abms{32.82}{1.29} & \abms{92.82}{1.00}
& \abms{68.28}{2.80} & \abms{60.63}{1.73} \\

& No Patch-Level SIGReg
& 31.34 & 91.45 & 67.09 & 46.03
& \abms{32.15}{1.66} & \abms{92.17}{0.63}
& \abms{67.69}{2.84} & \abms{54.29}{2.17} \\

& IDM only
& 30.74 & 91.71 & 65.48 & 46.41
& \abms{30.75}{1.73} & \abms{91.67}{1.04}
& \abms{65.90}{2.93} & \abms{58.12}{2.38} \\

& No IDM+FDM
& 31.29 & 91.66 & 66.41 & 45.56
& \abms{30.79}{1.21} & \abms{90.87}{0.82}
& \abms{65.12}{2.61} & \abms{55.09}{1.84} \\
\midrule

\multirow{4}{*}{DINOv2-L}
& Full
& 33.13 & 94.60 & 68.91 & 48.81
& \abms{38.14}{2.61} & \abms{96.03}{2.19}
& \abms{73.01}{2.98} & \abms{65.21}{2.92} \\

&  No Patch-Level SIGReg
& 31.54 & 91.69 & 66.90 & 43.08
& \abms{37.10}{2.87} & \abms{95.66}{0.39}
& \abms{71.05}{1.84} & \abms{64.19}{3.39} \\

& IDM only
& 30.89 & 94.35 & 66.87 & 47.81
& \abms{37.21}{2.62} & \abms{95.62}{1.95}
& \abms{72.02}{2.61} & \abms{64.22}{3.62} \\

& No IDM+FDM
& 33.52 & 94.26 & 67.26 & 48.09
& \abms{36.27}{1.88} & \abms{95.03}{1.53}
& \abms{71.21}{2.16} & \abms{63.77}{2.98} \\
\midrule

\multirow{4}{*}{SurgeNet-L}
& Full
& 32.78 & 95.02 & 71.64 & 50.89
& \abms{38.85}{2.07} & \abms{96.74}{1.24}
& \abms{72.82}{1.32} & \abms{65.73}{2.38} \\

&  No Patch-Level SIGReg
& 32.53 & 93.93 & 69.03 & 49.79
& \abms{38.17}{2.43} & \abms{96.19}{1.47}
& \abms{72.74}{2.98} & \abms{64.35}{3.65} \\

& IDM only
& 32.56 & 94.61 & 69.56 & 48.34
& \abms{37.60}{2.52} & \abms{96.11}{2.30}
& \abms{72.21}{3.20} & \abms{64.87}{3.00} \\

& No IDM+FDM
& 33.09 & 94.29 & 69.62 & 48.58
& \abms{34.81}{2.12} & \abms{96.42}{1.30}
& \abms{70.69}{1.90} & \abms{64.41}{2.85} \\
\bottomrule
\end{tabular}%
}
\end{table*}
\subsection{Main Recognition Performance and Ablations}
\label{sec:recognition}

% All methods trained under our \textbf{controlled protocol} use the same
% learning rate, fixed training seed, and number of epochs within
% each dataset. Published results marked with $\dagger$ are included
% for reference. CholecT50 results are obtained with one training seed,
% while ProstaTD results report mean $\pm$ standard deviation across
% five folds, using the same training seed in each fold. Unless otherwise stated,
% within each backbone, the ablations retain the same eight-frame
% input, temporal aggregator, and semantic objective, with $d_a=128$
% for all variants containing the IDM.

Table~\ref{tab:main_results} reports recognition on CholecT50 and ProstaTD.
With ViT-L, \ours{} reaches $32.15/32.82$ $\mathrm{mAP}_{IVT}$ on
CholecT50/ProstaTD, exceeding V-JEPA2 on CholecT50 ($29.68$) and
remaining comparable on ProstaTD ($32.66$). It also achieves higher
$\mathrm{mAP}_I$ on both datasets and comparable $\mathrm{mAP}_V/\mathrm{mAP}_T$
on CholecT50, showing that an image-pretrained ViT without temporal
pretraining can compete with a pretrained video encoder.
With DINOv2-L and SurgeNet-L, \ours{} reaches $38.14$ and $38.85$
$\mathrm{mAP}_{IVT}$ on ProstaTD, versus $35.87$ for V-JEPA2.1.
SurgeNet-L attains the highest mean triplet, instrument, and target mAP
among the listed methods, while V-JEPA2.1 retains the highest verb mAP.
On CholecT50, SurgeNet-L exceeds V-JEPA2.1 in all three component metrics
but not $\mathrm{mAP}_{IVT}$. Patch-level SIGReg has a relatively modest effect for
SurgeNet-L on ProstaTD. Section~\ref{sec:enc-behavior} studies the effects
of $d_a$ and prediction weight.

\paragraph{Component ablations at $d_a=128$.}

\begin{wraptable}{r}{0.45\linewidth}
\vspace{-8pt}
\centering
\caption{Relative mAP drop (\%) under INT4 quantization on CholecT50 ($d_a=128$); lower is better.}
\label{tab:deployment}

\small
\renewcommand{\arraystretch}{0.90}
\setlength{\tabcolsep}{3pt}
\setlength{\aboverulesep}{1pt}
\setlength{\belowrulesep}{1pt}

\resizebox{\linewidth}{!}{%
\begin{tabular}{@{}llrrrr@{}}
\toprule
Backbone & Variant
& $\Delta_{\mathrm{IVT}}$ & $\Delta_{\mathrm{I}}$
& $\Delta_{\mathrm{V}}$ & $\Delta_{\mathrm{T}}$ \\
\midrule

\multirow{3}{*}{ViT-L}
& Full       & \textbf{2.7} & 0.9 & 0.8 & 0.0 \\
& IDM only   & 10.5 & 1.9 & 3.5 & 0.6 \\
& No IDM+FDM & 4.2  & 0.1 & 0.5 & 2.9 \\
\midrule

\multirow{3}{*}{DINOv2-L}
& Full       & \textbf{2.8} & 0.3 & 0.2 & 0.1 \\
& IDM only   & 6.8  & 0.4 & 0.3 & 1.3 \\
& No IDM+FDM & 4.2  & 0.5 & 0.9 & 0.9 \\
\midrule

\multirow{3}{*}{SurgeNet-L}
& Full       & \textbf{$-$0.4} & 0.1 & 0.6 & $-$0.2 \\
& IDM only   & 0.3  & 0.0 & $-$1.4 & $-$0.6 \\
& No IDM+FDM & 1.3  & 0.2 & 1.0 & 0.5 \\
\midrule

V-JEPA2
& Baseline
& 7.7 & $-$0.2 & 2.4 & 3.9 \\

V-JEPA2.1
& Baseline
& 5.6 & 2.4 & 4.8 & 4.7 \\

\bottomrule
\end{tabular}%
}
\vspace{-8pt}
\end{wraptable}

Table~\ref{tab:combined_ablation} reports ablations at $d_a=128$.
Full ViT-L improves all four metrics on both datasets, gaining
$0.86/2.03$ $\mathrm{mAP}_{IVT}$ on CholecT50/ProstaTD.
DINOv2-L and SurgeNet-L improve verb/target mAP on CholecT50 but
remain slightly below their triplet baselines, while on ProstaTD all
three full models improve triplet, verb, and target mAP; DINOv2-L and
SurgeNet-L gain $1.87/4.04$ triplet points.
Adding FDM prediction and patch-level SIGReg over IDM alone further
improves triplet, verb, and target mAP across backbones and datasets.
Section~\ref{sec:enc-behavior} studies capacity and prediction weight. For temporal-direction sensitivity, full \ours{} increases verb reversal
drops from $13.8\!\to\!17.8\%$ (DINOv2-L),
$15.1\!\to\!18.5\%$ (ViT-L), and $19.0\!\to\!21.3\%$ (SurgeNet-L),
exceeding V-JEPA2/2.1 without large-scale video pretraining
(Table~\ref{tab:reversal_drop}). ProstaTD results are in Appendix~A.6.

% ============================================================
% COMPACT ABLATION
% ============================================================

\begin{wraptable}{r}{0.45\linewidth}
\vspace{-8pt}
\centering
\caption{
Relative mAP drop (\%) under clip reversal on CholecT50
($d_a=128$).
}
\label{tab:reversal_drop}

\small
\renewcommand{\arraystretch}{0.90}
\setlength{\tabcolsep}{3pt}
\setlength{\aboverulesep}{1pt}
\setlength{\belowrulesep}{1pt}

\resizebox{\linewidth}{!}{%
\begin{tabular}{@{}llrrrr@{}}
\toprule
Backbone & Variant
& $\Delta_{\mathrm{IVT}}$ & $\Delta_{\mathrm{I}}$
& $\Delta_{\mathrm{V}}$ & $\Delta_{\mathrm{T}}$ \\
\midrule

\multirow{3}{*}{ViT-L}
& Full       & 25.3 & 15.7 & 18.5 & 9.7 \\
& IDM only   & 21.4 & 14.1 & 16.0 & 9.2 \\
& No IDM+FDM & 21.8 & 13.9 & 15.1 & 13.0 \\
\midrule

\multirow{3}{*}{DINOv2-L}
& Full
& 27.9
& 16.6
& 17.8
& \underline{16.7} \\
& IDM only   & 23.1 & 15.7 & 16.4 & 14.4 \\
& No IDM+FDM & 19.6 & 13.1 & 13.8 & 16.1 \\
\midrule

\multirow{3}{*}{SurgeNet-L}
& Full
& \textbf{32.1}
& \textbf{18.9}
& \textbf{21.3}
& 14.7 \\
& IDM only
& 28.7
& \underline{18.6}
& \textbf{21.3}
& 14.2 \\
& No IDM+FDM
& 28.2
& 15.5
& \underline{19.0}
& 14.7 \\
\midrule

V-JEPA2
& Baseline
& 28.5
& 14.0
& 16.5
& \textbf{17.2} \\

V-JEPA2.1
& Baseline
& \underline{30.0}
& 17.7
& 17.0
& 14.4 \\

TimeSformer
& Baseline
& 5.1
& 3.8
& 3.4
& 4.2 \\

\bottomrule
\end{tabular}%
}
\vspace{-8pt}
\end{wraptable}

% ============================================================
% TEMPORAL REVERSAL
% ============================================================

% Requires \usepackage{booktabs}
\paragraph{Inference efficiency and deployment potential.}

For $T$ frames with $N$ patches, LAViF uses
$O(TN^2+T^2)$ attention versus $O(T^2N^2)$ for joint
spatiotemporal attention, and removes the FDM at inference.
ViT-L/DINOv2-L reach 72.8/50.8 clips/s, respectively. Under INT4 weight-only quantization of all linear layers, including the vision
encoder, IDM, aggregator, and text encoder, the full LAViF models lose at most $2.8\%$ in
$\mathrm{mAP}_{IVT}$ under INT4 quantization, compared with
1.3--4.2\% for the corresponding no-IDM+FDM baselines
and $5.6$--$7.7\%$ for V-JEPA2 and V-JEPA2.1 (Table~\ref{tab:deployment}).
We interpret this result as being consistent with greater low-bit tolerance
from decoupled frame-wise spatial encoding and compact temporal aggregation. More comprehensive results tables are provided in Appendix A.10.

% This component-specific response is consistent with target
% recognition relying more on appearance cues, whereas verb
% recognition benefits from distinguishing temporal order.

% \paragraph{Efficiency and quantization robustness.}

% Tables~\ref{tab:efficiency} and~\ref{tab:quant} evaluate inference
% cost and accuracy retention under reduced weight precision.
% Under the tested single-GPU configurations and input resolutions,
% \ours{} achieves $72.8$, $50.8$, and $24.1$ clips/s with
% ViT-L, DINOv2-L, and SurgeNet-L, respectively, compared with
% $10.4$--$14.7$ clips/s for the video baselines.
% ViT-L and DINOv2-L also use less peak GPU memory
% ($7.5$ and $10.9$,GB versus $15.4$--$22.0$,GB).
% For post-training weight-only quantization, INT8 changes
% $\mathrm{mAP}_{IVT}$ by less than $0.1$ points across all
% evaluated models.
% Under INT4, the changes are $-0.97$, $-0.82$, and $+0.11$
% points for our three backbones, compared with $-2.24$ and
% $-1.81$ points for V-JEPA2 and V-JEPA2.1.
% These measurements demonstrate faster inference and smaller
% INT4 accuracy changes for our evaluated configurations,
% supporting their potential for deployment with limited
% compute and memory.

\begin{figure}[t]
    \centering
    \includegraphics[width=\linewidth]{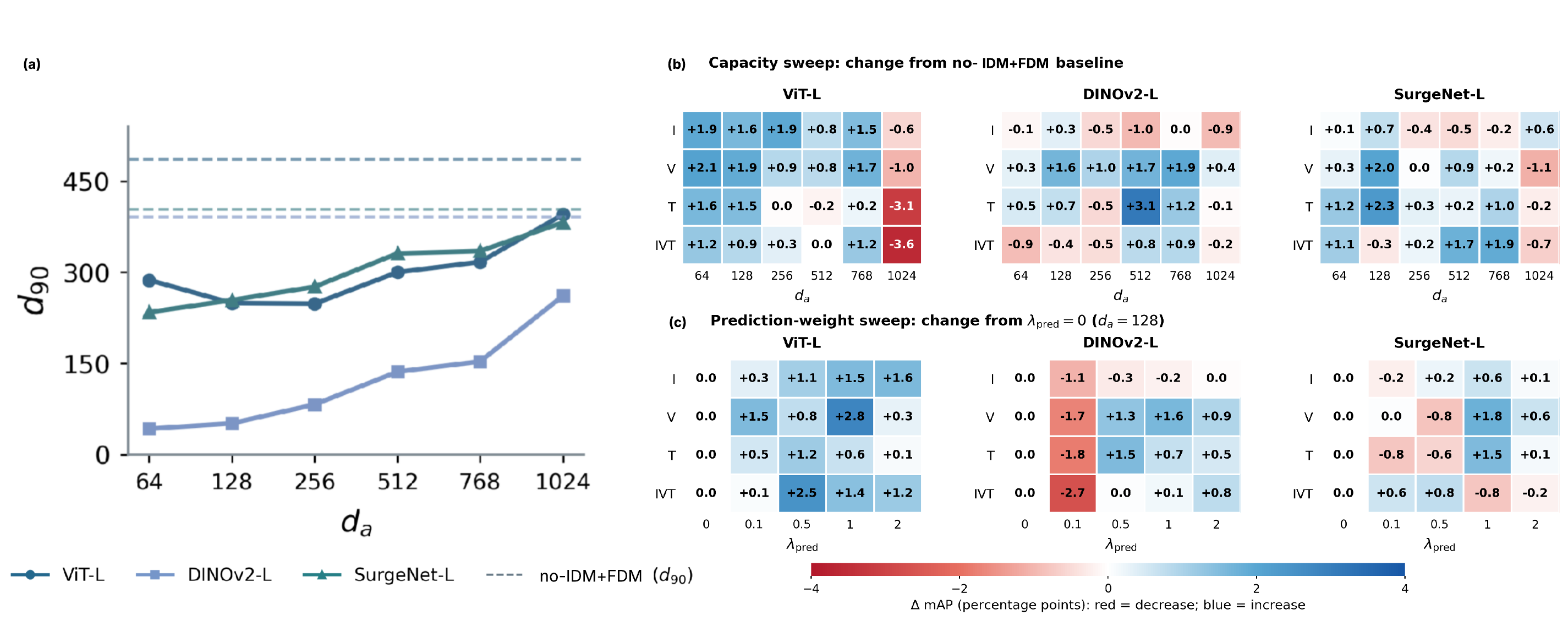}
    \vspace{-5mm}
    \caption{
\textbf{Transition structure and recognition under action-capacity and prediction-weight sweeps on CholecT50.}
\textbf{(a)} $d_{90}$ versus $d_a$; dashed lines denote no-IDM+FDM baselines.
\textbf{(b)} Recognition change across $d_a$ relative to the no-IDM+FDM baseline.
\textbf{(c)} Recognition change across $\lambda_{\mathrm{pred}}$ at $d_a=128$, relative to $\lambda_{\mathrm{pred}}=0$.
Heatmaps report I/V/T/IVT mAP differences (percentage points); blue/red indicate gains/drops.
}
\vspace{-5mm}
    \label{fig:theo_all}
\end{figure}

\begin{figure*}[t]
    \centering

    % (a) Main PCA-RGB comparison
    \begin{subfigure}[c]{0.85\textwidth}
        \centering
        \includegraphics[width=\linewidth]{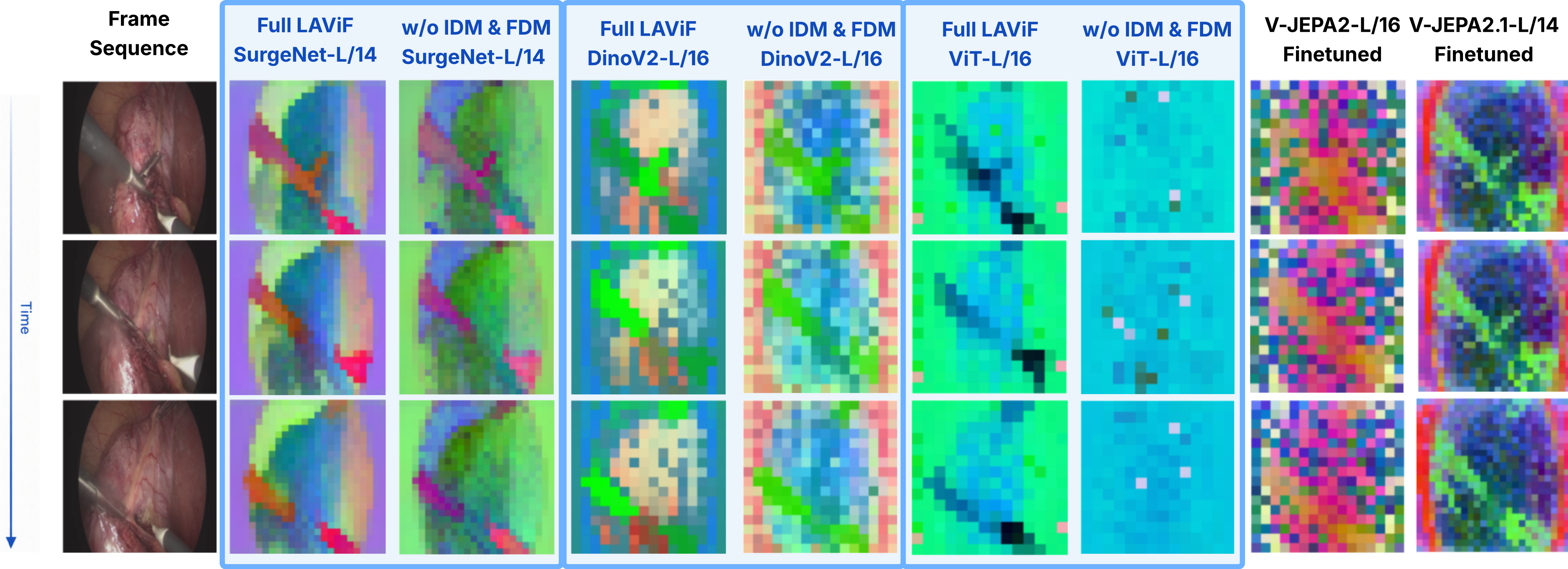}
        \caption{}
        \label{fig:all_pca_rgb}
    \end{subfigure}

    \vspace{-3pt}

    % (b) IoU bands
    \begin{subfigure}[c]{0.37\textwidth}
        \centering
        \includegraphics[width=\linewidth]{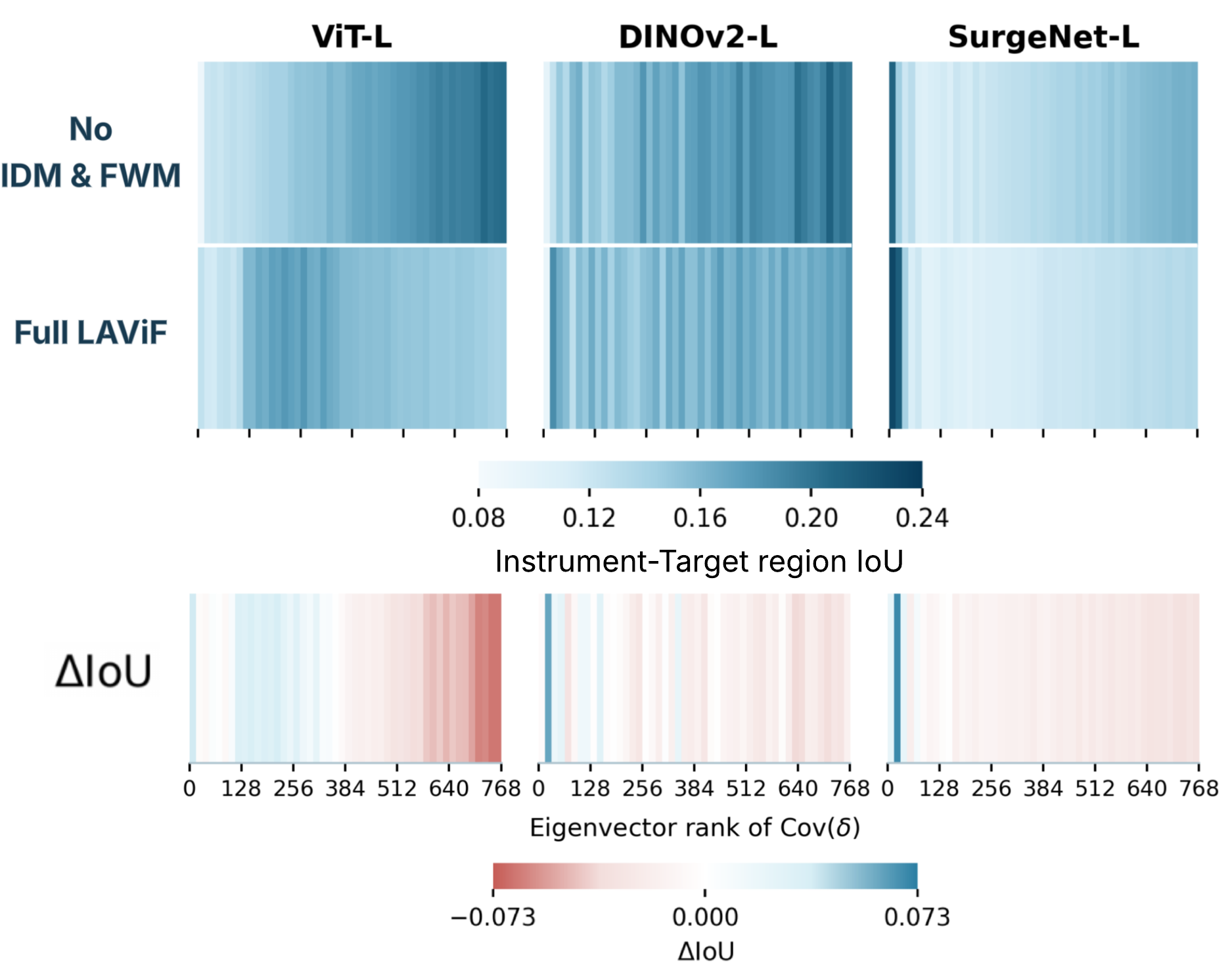}
        \caption{}
        \label{IoU}
    \end{subfigure}
    \hfill
    % (c) lambda_pred / SIGReg PCA-RGB
    \begin{subfigure}[c]{0.62\textwidth}
        \centering
        \includegraphics[width=\linewidth]{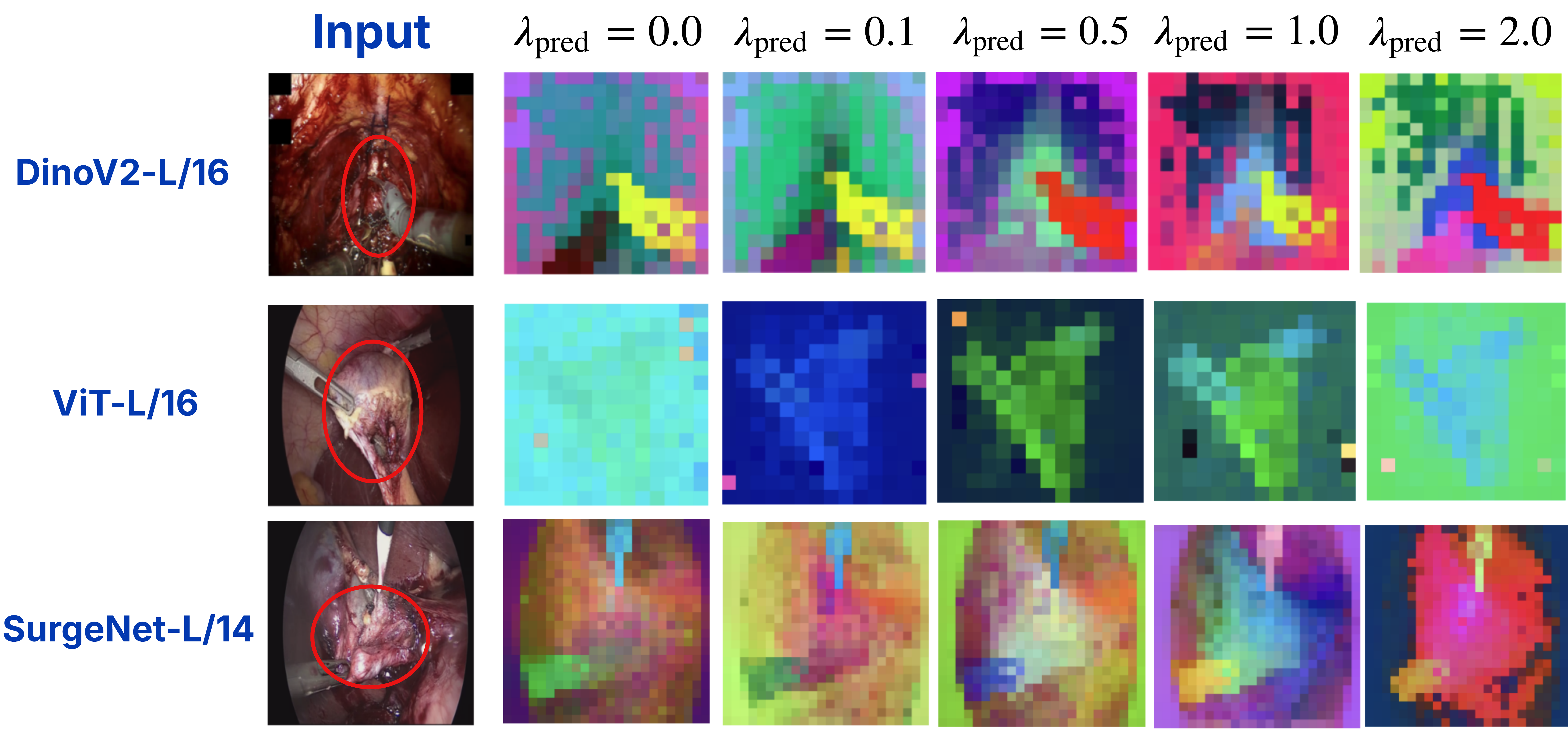}
        \caption{}
        \label{fig:pca_overlay}
    \end{subfigure}

    \caption{
\textbf{Spatial feature organization and interaction grounding.}
\textbf{(a)} PCA-RGB maps comparing \ours{}, no-IDM+FDM baselines, and V-JEPA2/2.1.
\textbf{(b)} Transition-band interaction-region IoU at $d_a=128$; $\Delta\mathrm{IoU}$ is relative to the no-IDM+FDM baseline.
\textbf{(c)} PCA-RGB maps across $\lambda_{\mathrm{pred}}$.
}
\vspace{-5mm}
    \label{fig:spatial_analysis}
\end{figure*}

\subsection{Transition Structure and Recognition}
\label{sec:enc-behavior}

\paragraph{Action capacity and predictive pressure.}
Under a simplified linear model~\citep{zhang2026latent},
the optimal prediction loss is
$\mathcal{L}_{\mathrm{pred}}^{*}=\sum_{i>d_a}\lambda_i$,
where $\lambda_i$ are transition-covariance eigenvalues in descending
order. Thus, $d_a$ controls how much dominant transition variance can
be represented by the latent action; smaller $d_a$ leaves more variance
outside the bottleneck and contributes to prediction error. In joint
training, this pressure may discourage task-irrelevant changes, while
semantic supervision preserves directions useful for recognition.
We summarize the resulting transition space using $d_{90}$, the number
of directions explaining $90\%$ of its variance. Figure~\ref{fig:theo_all}(a,b) shows that ViT-L improves all four
recognition metrics at $d_a\in\{64,128\}$ even though $d_{90}>d_a$,
indicating that the full transition space need not fit inside the action
bottleneck. DINOv2-L reaches its highest triplet mAP at $d_a=768$,
while removing the bottleneck at $d_a=d_s=1024$ lowers triplet mAP
relative to $d_a=768$ for all three encoders. SurgeNet-L similarly
favors $d_a=128$ for verb mAP but $d_a=768$ for triplet mAP.
At fixed $d_a=128$, increasing $\lambda_{\mathrm{pred}}$ from $0$ to $1$
improves ViT-L verb/triplet mAP by $2.81/1.40$ points and
SurgeNet-L verb/target mAP by $1.75/1.51$ points, while reducing its
triplet mAP by $0.79$ points (Figure~\ref{fig:theo_all}(c)).
Overall, predictive pressure can help select useful transitions, but stronger
pressure or greater action capacity does not uniformly benefit all components.

\paragraph{Spatial feature reorganization and transition grounding.}
PCA-RGB qualitatively reflects patch-level spatial coherence, while
transition-band IoU measures where dominant feature changes are grounded.
Figure~\ref{fig:all_pca_rgb} shows that \ours{} often makes target tissues
more distinct than the no-IDM+FDM baselines and V-JEPA2/2.1, with the
strongest changes for ViT-L and weaker effects for SurgeNet-L.
Figure~\ref{fig:pca_overlay} further suggests an encoder-dependent effect of
$\lambda_{\mathrm{pred}}$ on boundary clarity and local detail.
Using the setup of Section~\ref{sec:transition_analysis},
Figure~\ref{IoU} shows that \ours{} increases IoU in several
higher-variance transition bands while reducing it in many later bands.
Thus, latent-action training reorganizes both local patch structure and the
rank-dependent grounding of transition features rather than uniformly
improving localization. Additional examples and sweeps are provided in
Appendices A.7 and A.8.

\section{Related Work}\label{sec: related work}
\textbf{Surgical action-triplet recognition}
models $\langle\text{instrument},\text{verb},\text{target}\rangle$ interactions
using attention, temporal modeling, multi-task learning, and vision--language
supervision~\citep{nwoye2020tripnet,nwoye2022rdv,
sharma2023rit,li2023mtfist,sharma2025fine,walimbe2025adaptation,xi2023chain}.
Recent methods also use instrument trajectories, explicit interaction guidance,
or pseudo-labels~\citep{cheng2026trajpred,11084985,sharma2023surgical}.
We instead study how semantic alignment reorganizes visual representations
without additional spatial or motion supervision. \textbf{Latent-action models}
infer hidden actions from state transitions using inverse and forward dynamics,
primarily for action discovery, control, and VLA learning
~\citep{edwards2019ilpo,cui2024dynamo,ye2025lapa,schmidt2024lapo,
bu2025univla,sun2026vla}.
Prior methods often operate on fixed visual features, while predictive work
studies how such objectives shape representations
~\citep{maes2026lewm,zhang2026latent}.
In contrast, \ours{} jointly optimizes the visual encoder, IDM, FDM, temporal
aggregator, and semantic objective to reshape visual transition structure
during video--language alignment.
Extended discussion is in Appendix~A.1.

\section{Conclusion and Future Work}
We analyze how semantic alignment reshapes visual transition structure and
find that greater transition-space expansion does not necessarily imply
stronger recognition. Guided by this observation, we introduce \ours{} to
study whether latent-action prediction can reshape transition structure for
surgical interaction recognition without additional spatial or motion
supervision. Across multiple image backbones and two surgical datasets, we
evaluate recognition, interaction grounding, temporal-direction sensitivity,
and their dependence on $d_a$ and $\lambda_{\mathrm{pred}}$. Results show
encoder- and dataset-dependent reorganization of dominant transition
directions and local feature structure. Future work will study encoder-aware
parameter selection, broader surgical and non-surgical settings, and larger
multi-procedure pretraining.

% \bibliographystyle{plain}
% \bibliography{main}

\putbib[main]
\end{bibunit}
% ---- Appendix part: proof of Proposition \ref{prop:equiv} ----
% Uses the same preamble macros (\da, \Lp) and amsthm's `proof` environment.
% Do NOT restate the proposition in a proposition environment here (it would
% duplicate the label); the proof references the main-text statement.
\clearpage
\appendix
\begin{bibunit}[plainnat]
\section{Appendix}
\subsection{Related Work}

\paragraph{Surgical Action Triplet Recognition}

The surgical action triplet $\langle\text{instrument},\text{verb},\text{target}\rangle$ was introduced  as a fine-grained formalism for surgical workflow modeling. A triplet is fundamentally an \emph{interaction}: it is defined not by static appearance but by a change of state and the agent that causes it. ~\cite{nwoye2020tripnet} first recognized triplets directly from video with Tripnet (CholecT40), later extended into the Rendezvous (RDV) attention model and the CholecT50 dataset~\cite{nwoye2022rdv}, with standardized splits and metrics fixed in~\cite{nwoye2022splits}. Subsequent methods improve temporal modeling (RiT~\cite{sharma2023rit}, MT-FiST~\cite{li2023mtfist}), multi-task distillation~\cite{gui2024mt4mtl}, and parameter efficiency~\cite{li2024lam}. However, these classifier-based approaches rely heavily on well-annotated datasets and optimize closed-set category predictions, providing limited semantic alignment and transferability compared with pretrained vision encoders and VLMs. Recent surgical action-recognition methods increasingly leverage textual
semantics for vision--language alignment or as auxiliary supervision for
task-specific classifiers~\cite{sharma2025fine,walimbe2025adaptation,xi2023chain}.
However, these approaches remain largely image-centric and focus primarily on
prediction performance, with limited analysis of how semantic supervision
reshapes the underlying visual representation or what representation structure
is most suitable for surgical interaction understanding. In this work, we study
this representation-level question by analyzing frame-to-frame feature
transitions under semantic alignment, including their dimensionality, spatial
grounding, and sensitivity to temporal direction. Some surgical approaches further incorporate explicit spatial or motion cues,
including instrument trajectories, instrument--tissue interaction guidance,
or interaction pseudo-labels
~\citep{cheng2026trajpred,11084985,sharma2023surgical}.
These signals can provide useful localization priors, but they typically depend
on additional detectors, tracking pipelines, or pseudo-label generation.
In surgical videos, such automatically estimated cues may become unreliable
under occlusion, rapid camera motion, tissue deformation, and ambiguous or
blurred anatomical boundaries.
This motivates studying whether spatially grounded and temporally sensitive
representations can instead be learned directly through semantic alignment,
without requiring additional spatial or motion supervision during training.

\paragraph{Inverse Dynamics and Latent World Models.}

Latent-action models infer hidden causes of observed transitions without
requiring action labels, typically combining an inverse dynamics model (IDM)
that compresses state changes into latent actions with a forward dynamics model
(FDM) that predicts future states conditioned on them. Prior work has primarily
used this formulation for action discovery, policy learning, and robot control
~\cite{edwards2019ilpo,cui2024dynamo,schmidt2024lapo,
bu2025univla,sun2026vla}. Early methods learned discrete latent actions:
ILPO~\cite{edwards2019ilpo} learned a discrete action space for policy learning,
LAPO~\cite{schmidt2024lapo} used vector quantization to recover latent action
structure, and Genie~\cite{bruce2024genie} scaled discrete latent-action
modeling to internet-scale video. Subsequent work studied robustness to visual
distractors~\cite{nikulin2025laom}, continuous latent actions
~\cite{liang2025clam}, and latent-action supervision for VLA policy learning
~\cite{bu2025univla,sun2026vla}.
Importantly, several VLA approaches learn latent actions over fixed pretrained
visual representations: LAPA~\cite{ye2025lapa} freezes its vision encoder
during latent pretraining, while UniVLA~\cite{bu2025univla} learns its latent
action model in a fixed DINOv2 feature space. Thus, their latent-action
objectives do not directly reshape the underlying visual representation.
DynaMo~\cite{cui2024dynamo}, in contrast, jointly trains the visual encoder,
IDM, and FDM end-to-end for dynamics-aware visuomotor pretraining, but does not
study adaptation of a pretrained visual encoder under semantic language
supervision.
More recent work has examined representation learning through predictive
objectives. LeWM~\cite{maes2026lewm} uses stable end-to-end joint-embedding
prediction with SIGReg and shows that prediction can encourage features to
capture physical properties such as position and velocity. Complementary
analysis~\cite{zhang2026latent} shows that, for fixed state representations, a
compact IDM bottleneck can preserve dominant action-induced transition
directions, although nuisance variation may also be retained. In contrast,
our pretrained vision encoder, IDM, FDM, temporal aggregator, and semantic
objective are optimized jointly. We therefore study how latent-action
prediction reshapes the spatial organization and transition structure of a
pretrained visual representation during semantic VLM adaptation.

\subsection{Baseline Details}
We benchmark encoders from both general- and surgical-domain pretraining
regimes. General-domain models include CLIP ViT-L/14
~\cite{radford2021learning}, trained through image--text contrastive learning;
DINOv2-L/16~\cite{oquab2023dinov2}, trained through self-distillation and
masked-patch prediction; V-JEPA\,2/2.1~\citep{assran2025vjepa2,mur2026v},
which learn video representations through masked latent prediction,
and TimeSformer~\citep{bertasius2021space},
which models video with divided spatial--temporal attention. Surgical-domain
models include EndoViT~\cite{batic2023whether}, pretrained through masked
autoencoding; SurgeNet~\cite{JASPERS2026103873}, pretrained on surgical videos
using a DINO-style self-distillation objective; SurgVLP
~\cite{yuan2025learning}, trained by contrastively aligning surgical clips
with transcribed narration; HecVLP~\cite{yuan2024hecvl}, which extends this
alignment across multiple temporal levels; PeskaVLP
~\cite{yuan2024procedure}, which incorporates enriched text and
procedure-aware video--text alignment; and LEMON~\cite{che2026lemon}, trained
through augmented knowledge distillation. Additionally, we compare against models explicitly trained for the CholecT50 action-recognition task. Rendezvous~\cite{nwoye2022rdv} is a task-specific classifier that uses instrument-guided spatial attention and transformer-based semantic attention to recognize and associate the instrument, verb, and target. Chain-of-Look~\cite{xi2023chain} decomposes triplet recognition into a sequence of verb-centric visual-prompting steps and uses BioMedLM~\cite{bolton2024biomedlm} to adapt the generated prompts to the surgical domain. MML-SurgAdapt~\cite{walimbe2025adaptation} fine-tunes CLIP to jointly perform surgical phase recognition, critical-view-of-safety assessment, and action-triplet recognition within a unified model.

We instantiate our method on three image encoders chosen to represent distinct
pretraining paradigms: \textbf{ViT-L/16}, a vanilla vision transformer serving as the
most basic encoder; \textbf{DINOv2-L/16}, a self-distillation encoder representative of
methods that produce dense patch features (general-domain); and \textbf{SurgeNet-L/14},
which builds on top of DINOv2's architecture but is pretrained on surgical
data, forming a controlled pair with DINOv2 that isolates general versus surgical
pretraining of the same backbone. For all encoders we selected, we fully fine-tune them for contrastive learning aligned to the text of action triplets, where image encoders use single frames, video encoders and \ours{} use the whole clip (8 frames).  
%\noindent\textbf{Our \ours{} fine-tuning method.} Among these encoders, we select a representative subset and train them within our method.
 We further train two reduced settings for an ablation study. First, a \emph{states-only} model that removes the IDM+FDM from \ours{}: the aggregator $A_\omega$
temporally fuses the per-frame state tokens. This is basically a CLIP4Clip-style~\cite{luo2021clip4clip} model; this isolates the contribution of the latent-action module. Second, we ablate the
\emph{Patch-level SIG Regularizer}, training our method with and without it to measure its effect. For a fair comparison, each encoder uses the same EmbeddingGemma text encoder~\cite{vera2025embeddinggemma}.

\subsection{Spatial Back-Projection of Transition Directions}
\label{app:transition_grounding}

Let $\{v_k\}_{k=1}^{D}$ denote the orthonormal eigenvectors of
$\Sigma_\delta$, ordered by decreasing eigenvalue. We partition them
into consecutive bands
$B_j=\{v_{16(j-1)+1},\ldots,v_{16j}\}$.
For each consecutive-frame pair $(t,t+1)$, the transition of patch $p$
is $\Delta\phi_p^t=\phi_p^{t+1}-\phi_p^t$.
Its normalized projection energy onto band $B_j$ is
\begin{equation}
s_p^{(j)}
=
\frac{\sum_{v\in B_j}(v^\top\Delta\phi_p^t)^2}
{\|\Delta\phi_p^t\|_2^2+\varepsilon},
\end{equation}
where $\varepsilon$ ensures numerical stability.
This score measures the fraction of the patch's transition energy
captured by the band. We reshape the scores to the patch grid and
retain the top $25\%$ of patches to obtain a binary saliency map
$M_t^{(j)}$.

For CholecT50, we use SASVi-estimated segmentation
masks~\cite{sivakumar2025sasvi} to construct the evaluation region.
Let $Q_t$ denote the pixel-level union of the instrument and target
masks at frame $t$. We combine both transition endpoints and
downsample the resulting mask:
\begin{equation}
R_t=\operatorname{Downsample}_{\mathrm{any}}(Q_t\cup Q_{t+1}),
\end{equation}
where a patch-grid cell is active if it contains at least one active
pixel. This construction covers the instrument--target region at both
endpoints of the transition. We then compute
\begin{equation}
\operatorname{IoU}(M_t^{(j)},R_t)
=
\frac{|M_t^{(j)}\cap R_t|}
{|M_t^{(j)}\cup R_t|}.
\end{equation}

For each of five random seeds, we estimate $\Sigma_\delta$ from
$20{,}000$ consecutive-frame pairs and evaluate grounding on
$5{,}000$ sampled frames, each associated with a consecutive-frame
transition. We first average IoU over evaluation frames within each
seed, then report the mean and standard deviation of these per-band
averages across seeds.

For ProstaTD, no reliable instrument or tissue segmentation is
available: the dataset provides neither pixel-level masks nor
target-tissue annotations, so the instrument--target region used above
cannot be constructed. We therefore fall back to \emph{instrument-only}
localization obtained automatically. We detect surgical instruments in
each frame with  SAM2~\cite{ravi2025sam}, yielding
a per-frame set of instrument bounding boxes (boxes only; no tissue).
To suppress spurious full-frame detections we discard any box covering
more than $60\%$ of the image area. Let $\hat{Q}_t$ denote the pixel
region covered by the retained instrument boxes at frame $t$; the
evaluation region is formed exactly as before, by unioning both
transition endpoints and downsampling to the patch grid,
\begin{equation}
R_t=\operatorname{Downsample}_{\mathrm{any}}(\hat{Q}_t\cup \hat{Q}_{t+1}).
\end{equation}
Because $R_t$ marks only the instrument (not the target tissue), the
ProstaTD grounding score measures alignment with the \emph{instrument}
region rather than the full instrument--target interaction region; all
other settings ($16$-direction bands, top-$25\%$ saliency,
$20{,}000$-pair covariance estimate, $5{,}000$ evaluation frames) are
identical to CholecT50.

\subsection{Additional Method Details}
\label{app:method_details}

This section provides the implementation details.

\subsubsection{Frame States and Transitions}
\label{app:states_transitions}

For frame $x_t$, the vision encoder produces patch tokens
\[
\mathbf{p}_t=(p_{t,1},\ldots,p_{t,N})\in\mathbb{R}^{N\times d_s}.
\]
Unless otherwise specified, the pooled frame state is obtained by mean pooling:
\[
s_t=\frac{1}{N}\sum_{k=1}^{N}p_{t,k}.
\]
We then construct consecutive transitions
$\delta_t=s_{t+1}-s_t$ for $t=0,\ldots,T-1$.
The patch tokens $p_{t,k}$ used by the patch-level regularizer are taken
directly from this frame-level encoder output.

\subsubsection{Causal Transition IDM}
\label{app:causal_idm}

The Causal Transition IDM is implemented as a Transformer operating on the
transition sequence $(\delta_0,\ldots,\delta_{T-1})$. Its attention mask is
causal: the entry at row $t$ and column $j$ is unmasked only when $j\leq t$.
Consequently, the output at step $t$ depends only on
$\delta_0,\ldots,\delta_t$. A linear projection maps the Transformer output to
the latent-action dimension:
\[
a_t=W_a h_t+b_a,\qquad
h_t=\big[I_\theta(\delta_0,\ldots,\delta_{T-1})\big]_t.
\]
The causal mask ensures that the latent action cannot use later transitions
during inference.

\subsubsection{Forward Dynamics Model}
\label{app:fdm}

The FDM receives projected state tokens and latent actions. It uses a
DiT-style Transformer with adaptive layer-normalization-zero conditioning
~\cite{peebles2023scalable,maes2026lewm}. The state sequence is processed with
a causal mask, so the prediction at time $t$ can access only
$s_0,\ldots,s_t$ and $a_0,\ldots,a_t$. The corresponding action $a_t$
conditions the Transformer blocks through adaptive normalization:
\[
\hat{s}_{t+1}
=
\big[F_\psi(s_0,\ldots,s_{T-1};a_0,\ldots,a_{T-1})\big]_t .
\]
We train the FDM with a stop-gradient next-state target:
\[
\mathcal{L}_{\mathrm{pred}}
=
\frac{1}{T}
\sum_{t=0}^{T-1}
\left\|
\hat{s}_{t+1}-\operatorname{sg}(s_{t+1})
\right\|_2^2 .
\]
The FDM and this prediction loss are used only during training and are removed
at inference.

\subsubsection{Aggregation and Inference}
\label{app:aggregation}

Before aggregation, frame states and latent actions are projected to a shared
feature width. We prepend a learnable $\texttt{[AGG]}$ token and process
\[
[\texttt{[AGG]},s_0,\ldots,s_T,a_0,\ldots,a_{T-1}]
\]
with a bidirectional Transformer. The output corresponding to
$\texttt{[AGG]}$ is used as the clip representation $v$. Unlike the IDM and
FDM, the aggregator is bidirectional because the complete input clip is
available for recognition. At inference, only the vision encoder, IDM, and
aggregator are executed; the FDM is not evaluated.

\subsubsection{Semantic Alignment and Recognition}
\label{app:semantic_alignment}

EmbeddingGemma~\cite{vera2025embeddinggemma} encodes each textual class
description into token representations, which are mean-pooled to form the
text embedding. For a triplet class $c$, let $g_c$ denote its text embedding.
The clip and text embeddings are converted to logits using temperature-scaled
cosine similarity:
\begin{equation}
z_c
=
\tau
\frac{\langle v,g_c\rangle}
{\|v\|_2\|g_c\|_2}
+b,
\end{equation}
where $\tau$ is a learnable scale and $b$ is a learnable bias.

We construct text descriptions for the complete
instrument--verb--target triplet and for each individual instrument, verb, and
target component. Component labels are obtained deterministically from the
fixed triplet-to-component mapping: a component is positive whenever it occurs
in any positive triplet. We use binary cross-entropy with logits for all
triplet and component predictions:
\begin{equation}
\mathcal{L}_{\mathrm{cls}}
=
\lambda_{\mathrm{ivt}}
\operatorname{BCEWithLogits}(z,y)
+
\sum_{q\in\{\mathrm{i},\mathrm{v},\mathrm{t}\}}
\lambda_q
\operatorname{BCEWithLogits}(z^q,y^q).
\end{equation}
Here, $(z,y)$ denotes the complete-triplet logits and labels, while
$(z^q,y^q)$ denotes the corresponding component logits and labels. We use
$\lambda_q=1.0$ for the instrument, verb, and target objectives. Component-level
semantic supervision provides complementary cues for fine-grained action
recognition~\cite{wray2019fine,momeni2023verbs,vo2024henasy,
sharma2025fine}.

\subsubsection{SIG Regularization}
\label{app:sigreg}

We adopt SIGReg~\cite{maes2026lewm,balestriero2025lejepa}. For a set of
representations $Z=\{z_i\}_{i=1}^{N}$, SIGReg compares random
one-dimensional projections of the empirical representation distribution with a
standard Gaussian:
\begin{equation}
\operatorname{SIGReg}(Z)
=
\frac{1}{M}
\sum_{m=1}^{M}
\mathcal{T}
\left(
\left\{
\langle z_i,u_m\rangle
\right\}_{i=1}^{N}
\right),
\end{equation}
where $u_m$ is sampled from the unit sphere and
$\mathcal{T}(\cdot)$ is the closed-form sliced Cramér--Wold
Epps--Pulley statistic. Projection directions are resampled at every
optimization step.

For state/action regularization, pooled states and latent actions are collected
across the minibatch and regularized using $M=1024$ random projections. For
the patch-level term, in this work we uniformly sample $32$ patch tokens from each frame at each training step and
use $M=256$ projections:
\begin{equation}
\mathcal{L}_{\mathrm{SIG}}
=
\operatorname{SIGReg}(\{s_t\})
+
\operatorname{SIGReg}(\{a_t\}),
\end{equation}
\begin{equation}
\mathcal{L}_{\mathrm{SIG}}^{\mathrm{patch}}
=
\operatorname{SIGReg}
\left(
\{p_{t,k}:k\in\mathcal{K}_t\}
\right),
\qquad
|\mathcal{K}_t|=32 .
\end{equation}
The regularization contribution in the main objective is
\[
\mathcal{L}_{\mathrm{reg}}
=
\lambda_s\mathcal{L}_{\mathrm{SIG}}
+
\lambda_p\mathcal{L}_{\mathrm{SIG}}^{\mathrm{patch}}.
\]

\subsubsection{Optimization and Ablation Settings}
\label{app:optimization}

Unless otherwise specified, all controlled comparisons use the same backbone,
eight-frame input, temporal aggregator, semantic objective, training schedule,
and number of epochs. The full model uses
$\lambda_{\mathrm{pred}}=1.0$. Setting
$\lambda_{\mathrm{pred}}=0$ removes the prediction loss while retaining the
IDM and the state--action recognition pathway. Removing IDM+FDM and the
patch-level SIGReg produces the corresponding controlled baseline. Backbone-
specific dimensions, learning rates, loss weights, and optimization
hyperparameters are listed in Table~\ref{tab:lr}.

\subsection{Implementation Details}
\begin{table}[H]
\centering
\small
\setlength{\tabcolsep}{4pt}
\caption{Per-component learning rates for each vision encoder. \textbf{Encoder}
is the vision backbone's rate. Weight decay is shared across all groups.}
\label{tab:lr}
\begin{tabular}{lcccccc}
\toprule
\textbf{Experiments} & \textbf{Encoder} & \textbf{IDM} & \textbf{FDM}
& \textbf{Agg.} & \textbf{Text} & \textbf{Weight} \\
 & \textbf{LR} & \textbf{LR} & \textbf{LR}
& \textbf{LR} & \textbf{LR} & \textbf{decay} \\
\midrule
Ours ViT-L and action token dimension ablation
& $3{\times}10^{-5}$ & $3{\times}10^{-5}$ & $3{\times}10^{-5}$
& $3{\times}10^{-5}$ & $3{\times}10^{-5}$ & $0.01$ \\

Ours DINOv2 and action token dimension ablation
& $3{\times}10^{-5}$ & $1{\times}10^{-4}$ & $5{\times}10^{-5}$
& $1{\times}10^{-4}$ & $3{\times}10^{-5}$ & $0.01$ \\

Ours SurgeNet and action token dimension ablation
& $3{\times}10^{-5}$ & $5{\times}10^{-5}$ & $5{\times}10^{-5}$
& $1{\times}10^{-4}$ & $3{\times}10^{-5}$ & $0.01$ \\

No IDM\&FDM ViT-L Baseline
& $3{\times}10^{-5}$ & - & -
& $3{\times}10^{-5}$ & $3{\times}10^{-5}$ & $0.01$ \\

No IDM\&FDM DINOv2 Baseline
& $3{\times}10^{-5}$ & - & -
& $1{\times}10^{-4}$ & $3{\times}10^{-5}$ & $0.01$ \\

No IDM\&FDM SurgeNet Baseline
& $3{\times}10^{-5}$ & - & -
& $1{\times}10^{-4}$ & $3{\times}10^{-5}$ & $0.01$ \\

All other baselines
& $3{\times}10^{-5}$ & - & -
& - & $3{\times}10^{-5}$ & $0.01$ \\
\bottomrule
\end{tabular}
\end{table}

\subsection{Temporal-direction sensitivity.}

Table~\ref{tab:reversal_drop} shows that adding our Causal
Transition IDM increases verb reversal drops across all three
backbones. Adding FDM prediction with its accompanying patch-level
SIG regularization further increases verb sensitivity for DINOv2
and ViT-L, and triplet sensitivity for all three encoders.
Interestingly, mAP of Target reversal drops remain close to baseline for DINOv2
($16.1\%$ to $16.7\%$), remain unchanged for SurgeNet
($14.7\%$), and decrease for ViT-L ($13.0\%$ to $9.7\%$).

\subsection{Additional results on CholecT50 and ProstaTD}

\begin{figure*}[p]
    \centering

    \begin{subfigure}[t]{0.49\textwidth}
        \centering
        \includegraphics[width=\linewidth]{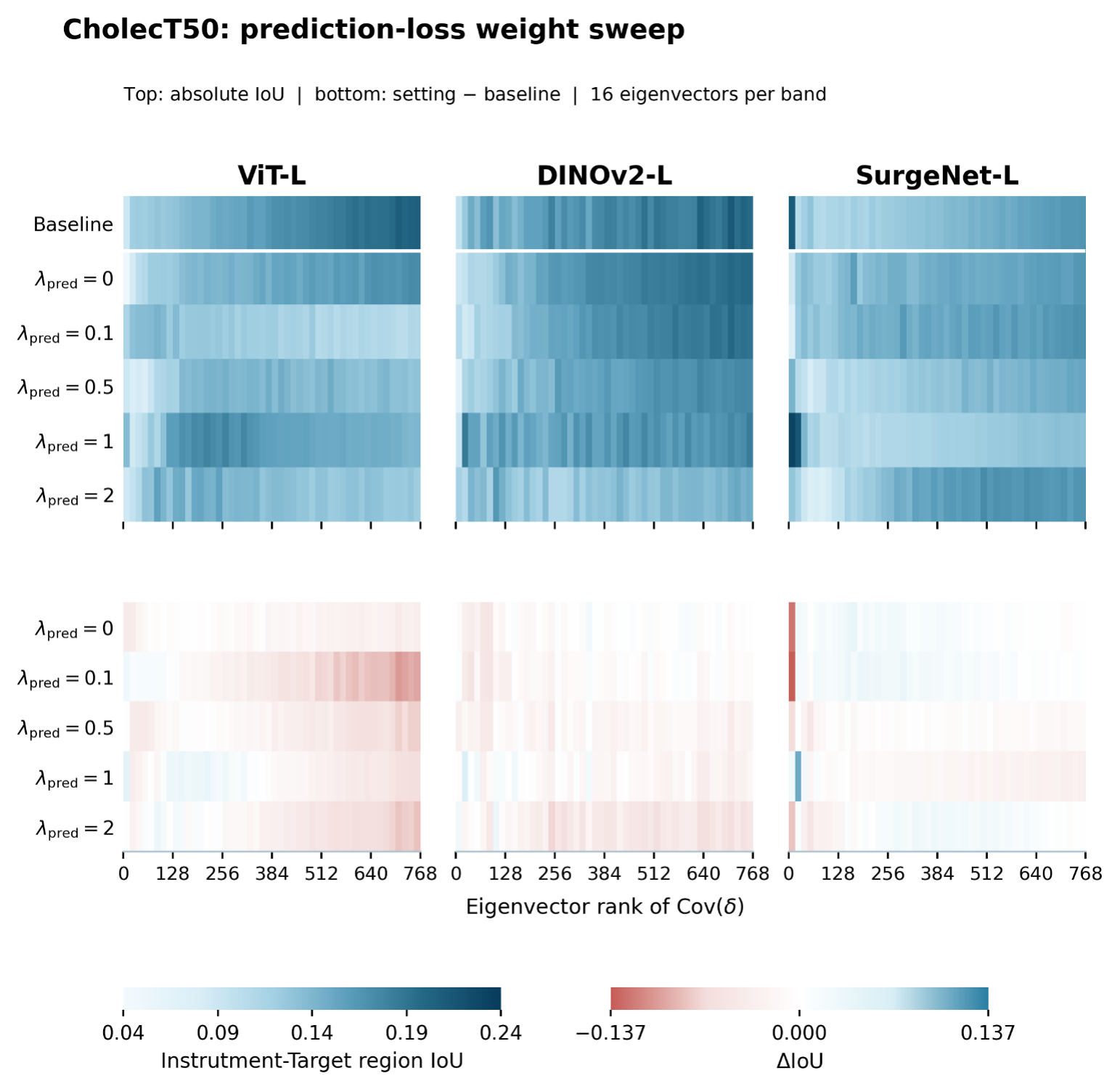}
        \caption{CholecT50: $\lambda_{\mathrm{pred}}$ sweep at $d_a=128$.}
        \label{fig:app_cholect50_iou_lambda}
    \end{subfigure}
    \hfill
    \begin{subfigure}[t]{0.49\textwidth}
        \centering
        \includegraphics[width=\linewidth]{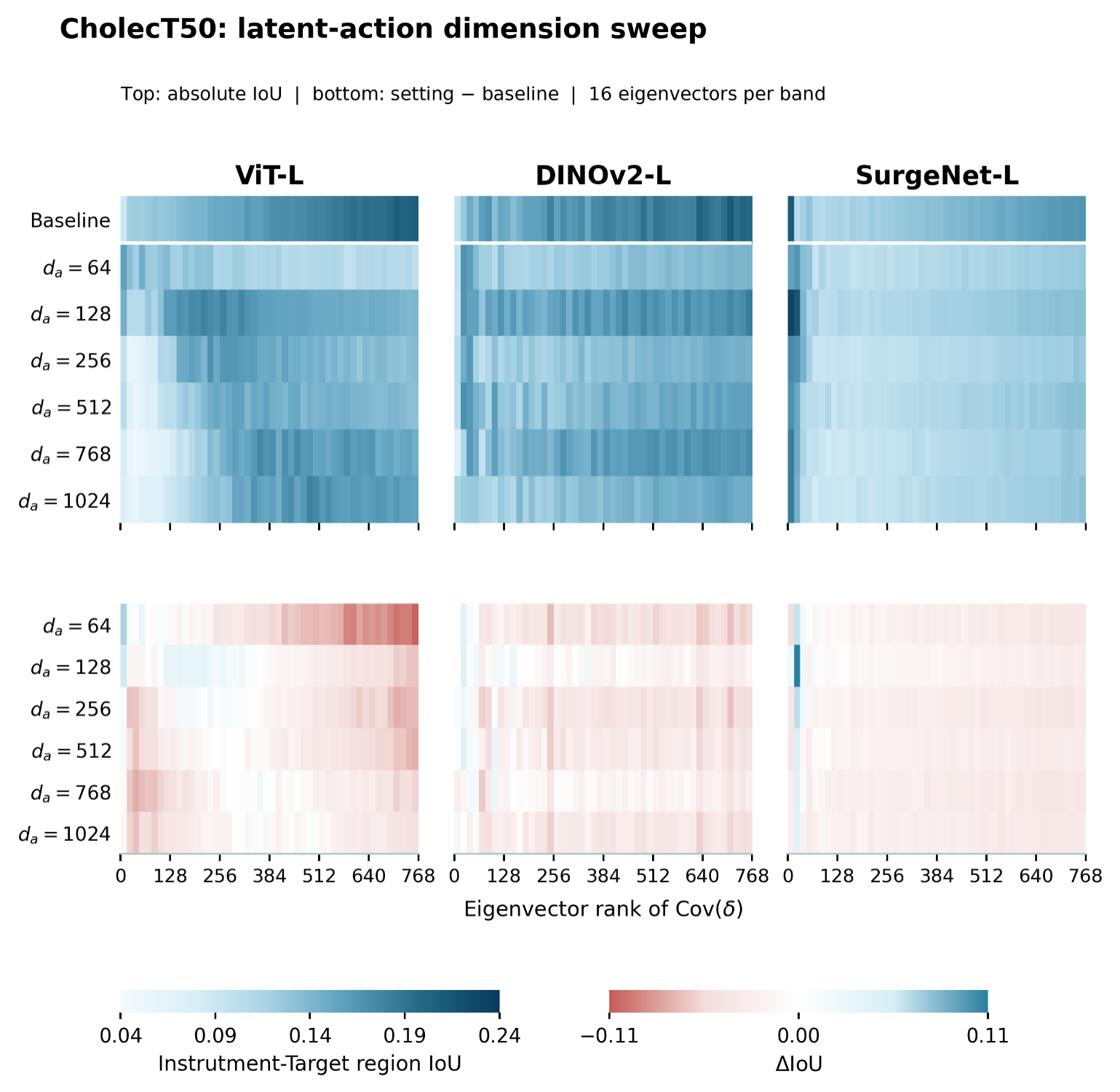}
        \caption{CholecT50: $d_a$ sweep.}
        \label{fig:app_cholect50_iou_da}
    \end{subfigure}

    \medskip

    \begin{subfigure}[t]{0.49\textwidth}
        \centering
        \includegraphics[width=\linewidth]{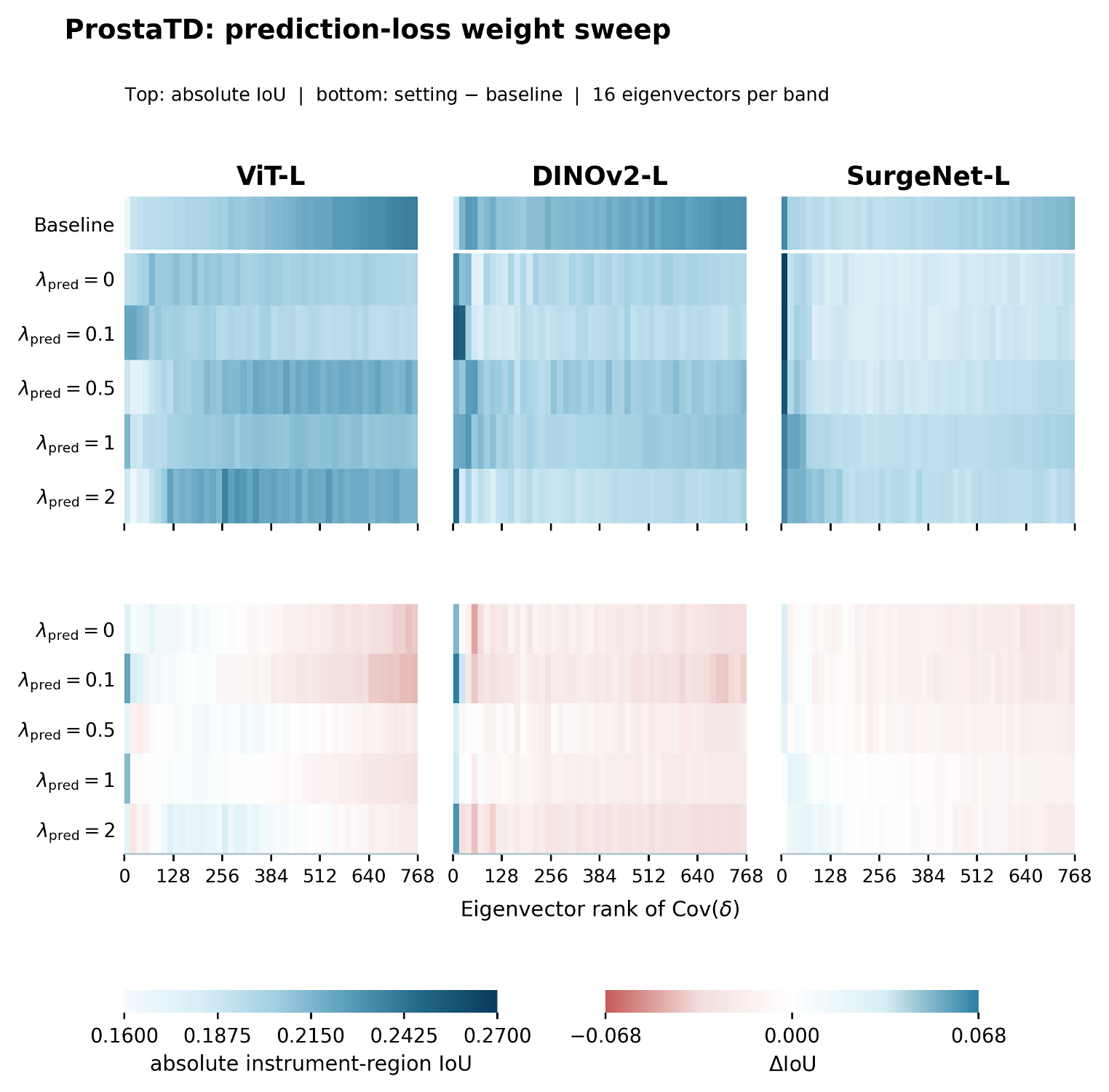}
        \caption{ProstaTD: $\lambda_{\mathrm{pred}}$ sweep at $d_a=128$.}
        \label{fig:app_prostatd_iou_lambda}
    \end{subfigure}
    \hfill
    \begin{subfigure}[t]{0.49\textwidth}
        \centering
        \includegraphics[width=\linewidth]{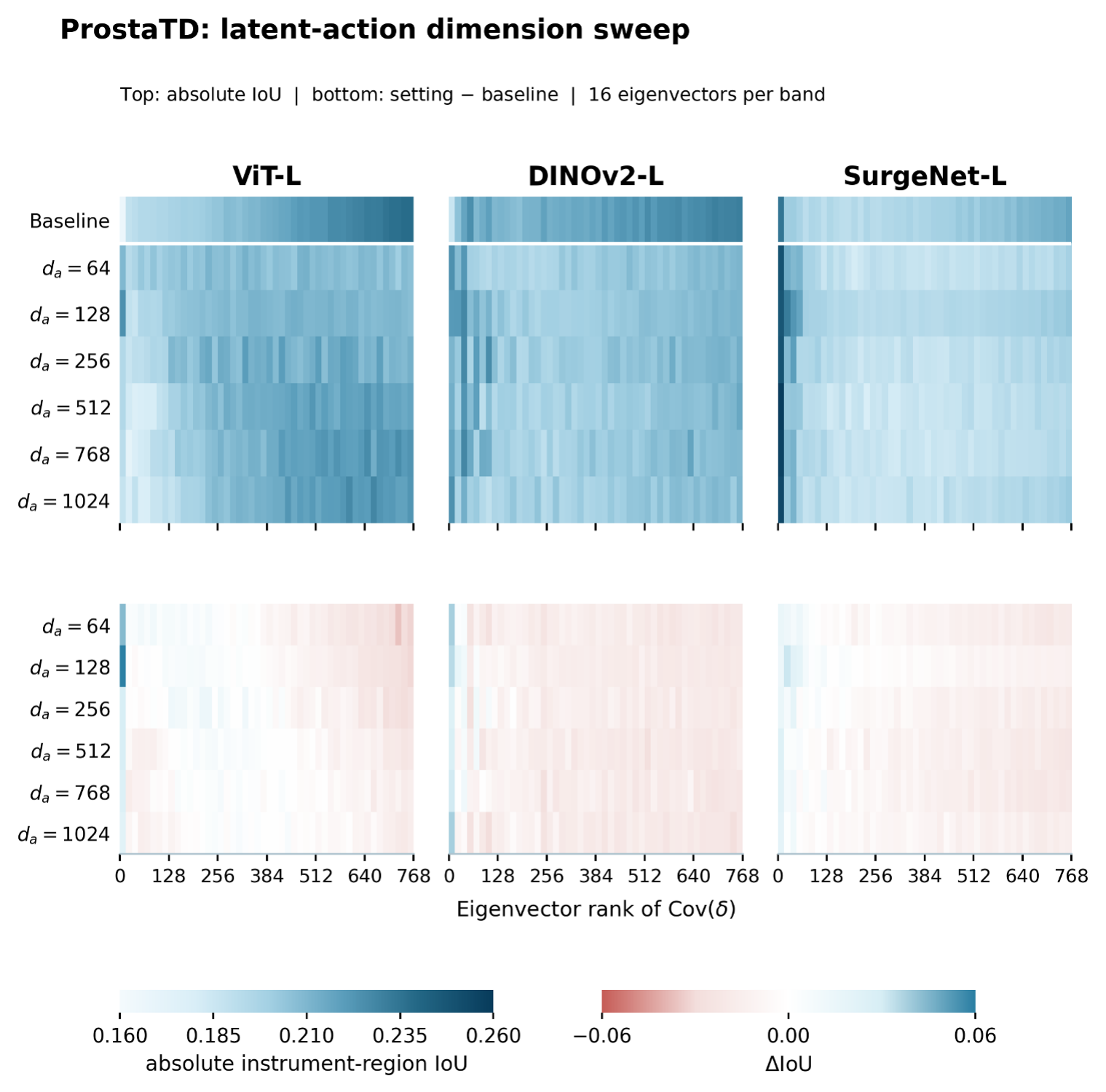}
        \caption{ProstaTD: $d_a$ sweep.}
        \label{fig:app_prostatd_iou_da}
    \end{subfigure}

    \caption{
    \textbf{Spatial grounding under prediction-weight and action-capacity sweeps.}
    Left: varying $\lambda_{\mathrm{pred}}$ at fixed $d_a=128$.
    Right: varying $d_a$.
    Transition eigen-bands are ordered from higher to lower variance.
    Absolute IoU measures overlap with the instrument--target region;
    $\Delta\mathrm{IoU}$ is the difference from the same encoder's
    no-IDM+FDM baseline.
    Blue/red in the difference maps indicate increased/decreased IoU,
    while white indicates little change. ProstaTD IoU uses SAM2-estimated instrument masks and should be
interpreted with caution due to potential mask inaccuracies caused by model domain gap.
    }
    \label{fig:app_iou_sweeps}
\end{figure*}

\paragraph{Spatial grounding under prediction-weight and action-capacity sweeps.}
We sweep $\lambda_{\mathrm{pred}}$ at a fixed $d_a=128$ on
CholecT50 and ProstaTD to examine how prediction strength affects
the spatial grounding of transition directions.
We separately vary $d_a$ to examine the effect of action capacity
(Figure~\ref{fig:app_iou_sweeps}).
For each encoder and eigen-band, $\Delta\mathrm{IoU}$ denotes
the IoU under the corresponding setting minus that of its
no-IDM+FDM baseline.
These analyses complement the recognition results by examining
how spatial grounding varies across transition ranks and
training configurations.
For ProstaTD, instrument bounding-box annotations were not available
to us at the time of evaluation, so we use SAM2~\cite{ravi2025sam} to estimate instrument
masks. Since SAM2 is not specifically trained for surgical scenes,
these masks may contain errors; the resulting IoU values should
therefore be interpreted with caution, particularly when comparing
across datasets.
% Preamble:
% \usepackage{graphicx}
% \usepackage{subcaption}

\begin{figure}[p]
    \centering
    \includegraphics[width=\linewidth]{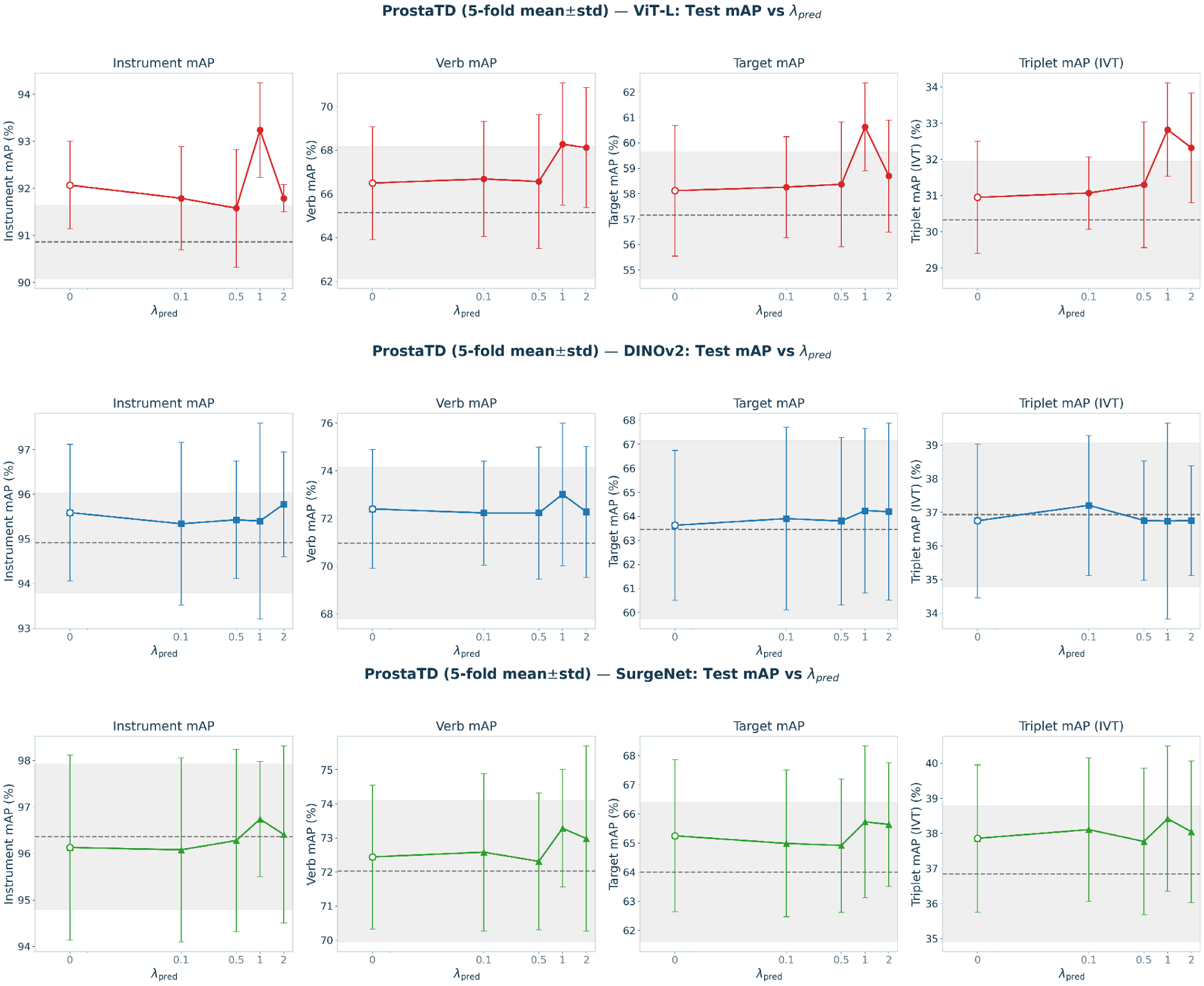}
    \caption{
    \textbf{Prediction-weight sweep on ProstaTD at fixed $d_a=128$.}
    Recognition performance of ViT-L, DINOv2-L, and SurgeNet-L
    under different $\lambda_{\mathrm{pred}}$ settings,
    reported as $\mathrm{mAP}_{I}$, $\mathrm{mAP}_{V}$,
    $\mathrm{mAP}_{T}$, and $\mathrm{mAP}_{IVT}$.
    }
    \label{fig:app_prostatd_lambda_recognition}
\end{figure}

\paragraph{Prediction-weight sweeps across action capacities and datasets.}
We extend the fixed-$d_a=128$ analysis by sweeping
$\lambda_{\mathrm{pred}}\in\{0,0.1,0.5,1,2\}$ at
$d_a\in\{64,128,512,1024\}$ on CholecT50
(Figure~\ref{fig:app_da_lambda_recognition}(a-c)).
Across ViT-L, DINOv2-L, and SurgeNet-L, the recognition response
to prediction supervision changes with action capacity and
triplet component.
These sweeps show that the response observed at $d_a=128$
does not fully characterize an encoder's behavior at other
capacities, motivating joint consideration of bottleneck size
and prediction weight.
We additionally sweep $\lambda_{\mathrm{pred}}$ on ProstaTD
at fixed $d_a=128$ to examine prediction-weight sensitivity
on a second surgical dataset
(Figure~\ref{fig:app_prostatd_lambda_recognition}).
% Preamble:
% \usepackage{graphicx}
% \usepackage{caption}

\begin{figure}[p]
    \centering
    \textbf{(a)}\\[-2pt]
    \includegraphics[width=\linewidth]{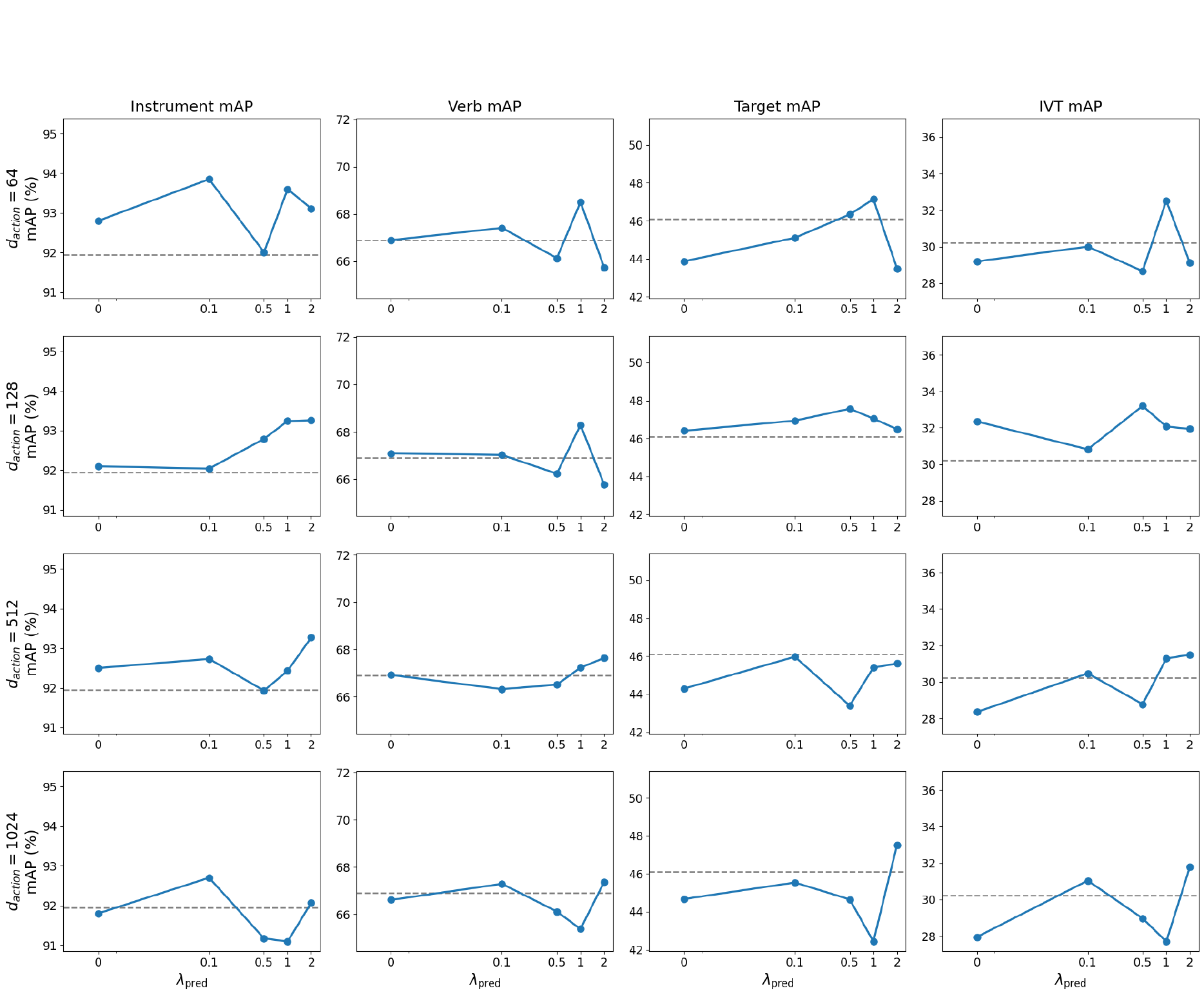}
    \caption{
    \textbf{Prediction-weight sweeps across action capacities on CholecT50: ViT-L.}
    Rows correspond to $d_a\in\{64,128,512,1024\}$;
    columns report $\mathrm{mAP}_{I}$, $\mathrm{mAP}_{V}$,
    $\mathrm{mAP}_{T}$, and $\mathrm{mAP}_{IVT}$.
    The horizontal axis varies $\lambda_{\mathrm{pred}}$,
    and dashed lines indicate the same-backbone no-IDM+FDM baseline.
    Each metric uses the same vertical scale across capacities and encoders.
    }
    \label{fig:app_da_lambda_recognition}
    \label{fig:app_vitl_da_lambda}
\end{figure}

\begin{figure}[p]
    \ContinuedFloat
    \centering
    \textbf{(b)}\\[-2pt]
    \includegraphics[width=\linewidth]{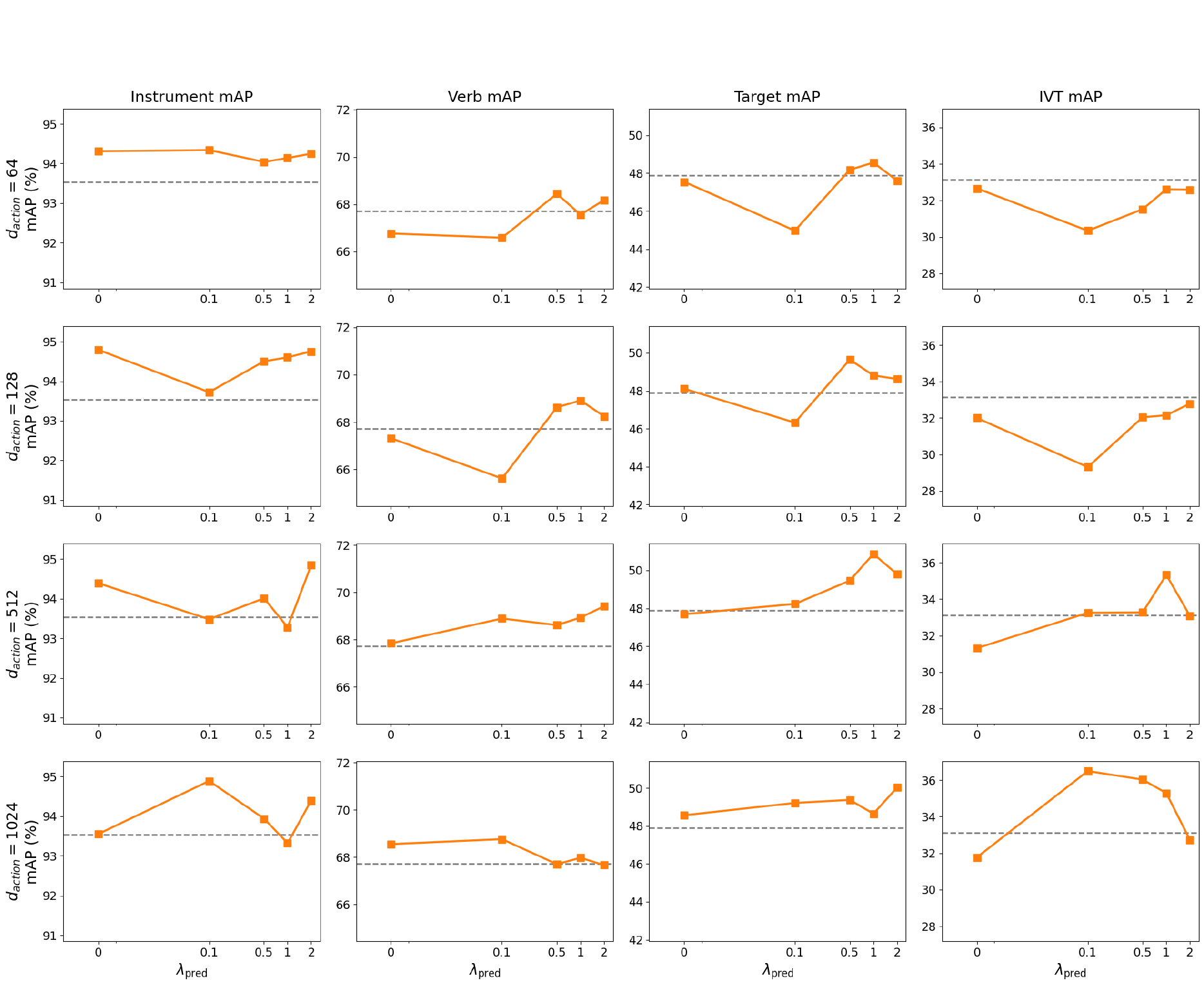}
    \caption[]{
    \textbf{Prediction-weight sweeps across action capacities on CholecT50: DINOv2-L.}
    Rows vary $d_a$; columns show instrument, verb, target, and triplet mAP.
    Dashed lines indicate the same-backbone no-IDM+FDM baseline.
    Axes and scales follow the ViT-L panel.
    }
    \label{fig:app_dinov2_da_lambda}
\end{figure}

\begin{figure}[p]
    \ContinuedFloat
    \centering
    \textbf{(c)}\\[-2pt]
    \includegraphics[width=\linewidth]{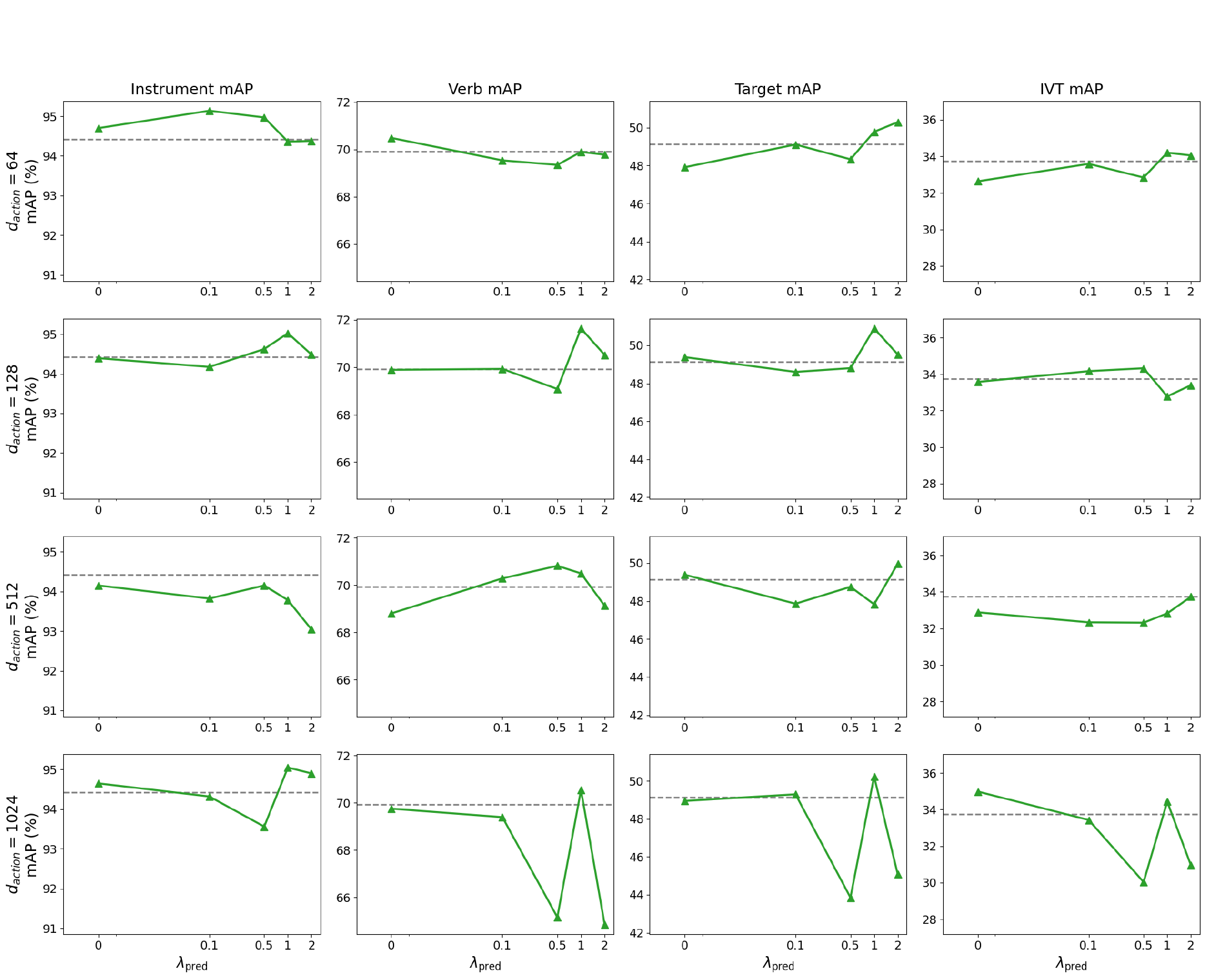}
    \caption[]{
    \textbf{Prediction-weight sweeps across action capacities on CholecT50: SurgeNet-L.}
    Rows vary $d_a$; columns show instrument, verb, target, and triplet mAP.
    Dashed lines indicate the same-backbone no-IDM+FDM baseline.
    Axes and scales follow the ViT-L panel.
    }
    \label{fig:app_surgenet_da_lambda}
\end{figure}

\begin{figure}[p]
    \centering
    \includegraphics[
        width=\linewidth,
        height=0.82\textheight,
        keepaspectratio
    ]{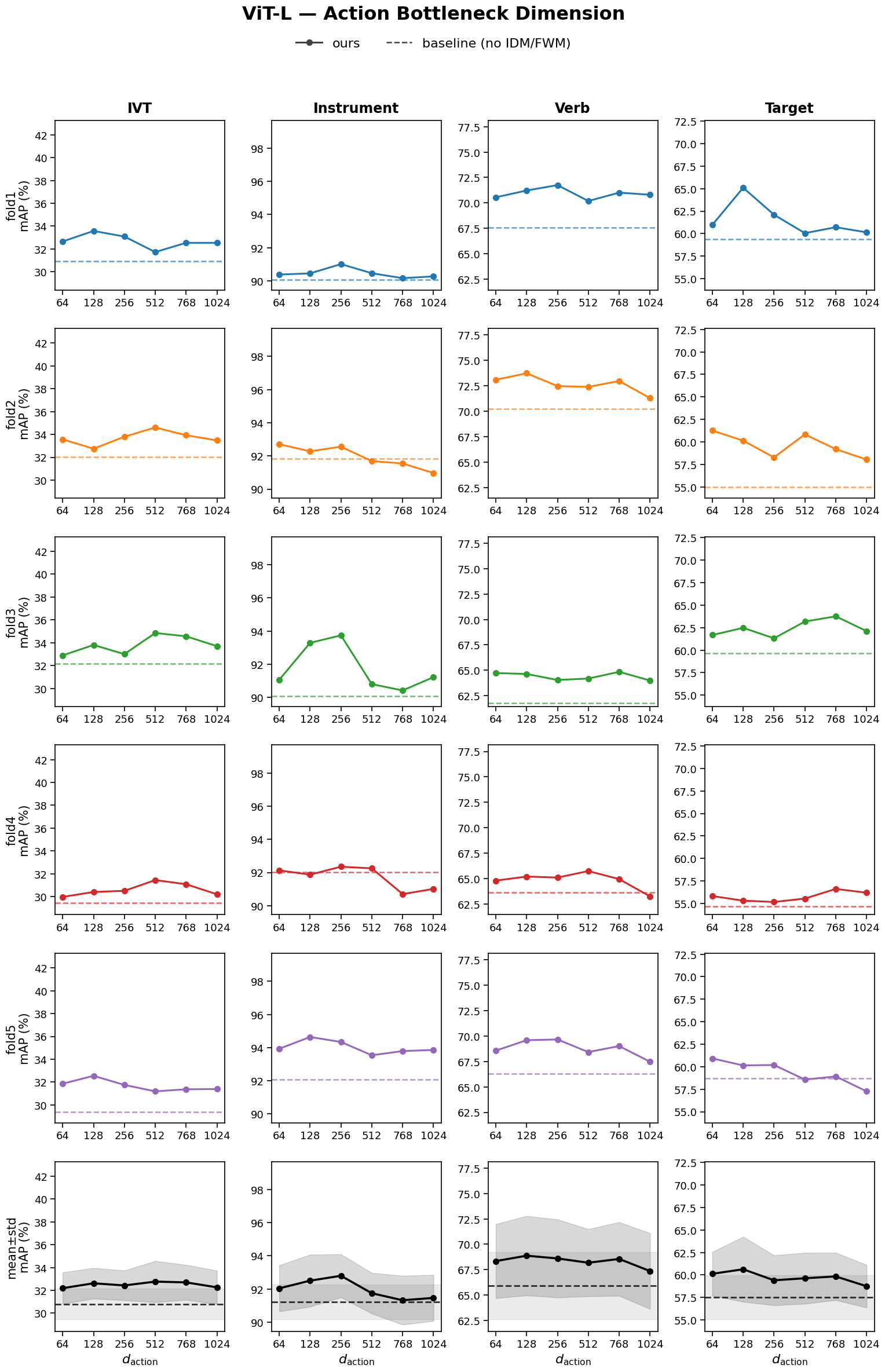}
    \caption{
    \textbf{Effect of latent-action dimension on ProstaTD: ViT-L.}
    Columns report test $\mathrm{mAP}_{IVT}$, $\mathrm{mAP}_{I}$,
    $\mathrm{mAP}_{V}$, and $\mathrm{mAP}_{T}$ across
    $d_a\in\{64,128,256,512,768,1024\}$.
    The first five rows show individual cross-validation folds;
    the bottom row shows their mean with shaded standard deviation.
    Dashed lines indicate the corresponding no-IDM+FDM baseline.
    }
    \label{fig:prostatd_daction_vitl}
\end{figure}

\begin{figure}[p]
    \centering
    \includegraphics[
        width=\linewidth,
        height=0.82\textheight,
        keepaspectratio
    ]{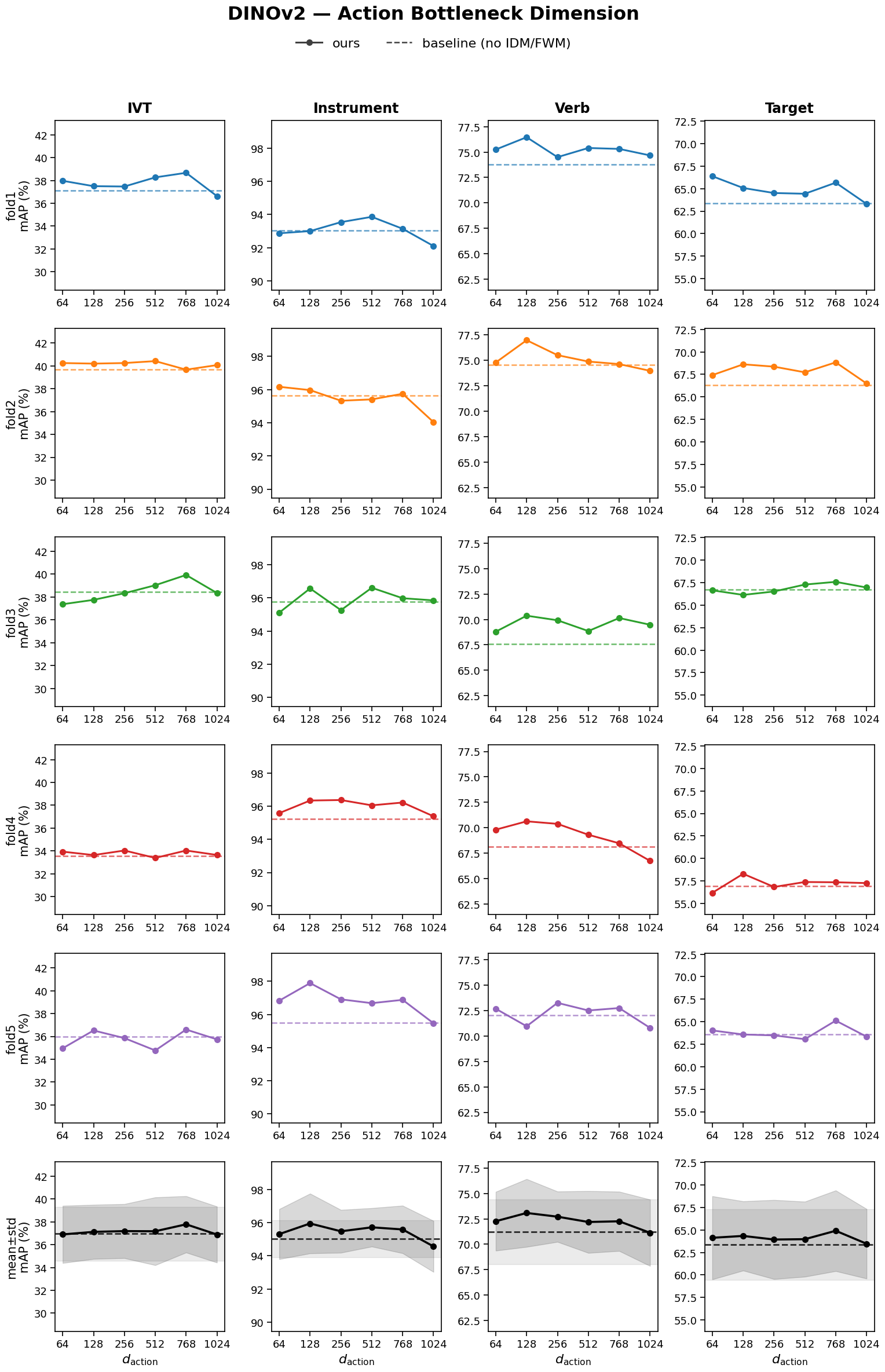}
    \caption{
    \textbf{Effect of latent-action dimension on ProstaTD: DINOv2-L.}
    Columns report test $\mathrm{mAP}_{IVT}$, $\mathrm{mAP}_{I}$,
    $\mathrm{mAP}_{V}$, and $\mathrm{mAP}_{T}$ across
    $d_a\in\{64,128,256,512,768,1024\}$.
    The first five rows show individual cross-validation folds;
    the bottom row shows their mean with shaded standard deviation.
    Dashed lines indicate the corresponding no-IDM+FDM baseline.
    }
    \label{fig:prostatd_daction_dinov2}
\end{figure}

\begin{figure}[p]
    \centering
    \includegraphics[
        width=\linewidth,
        height=0.82\textheight,
        keepaspectratio
    ]{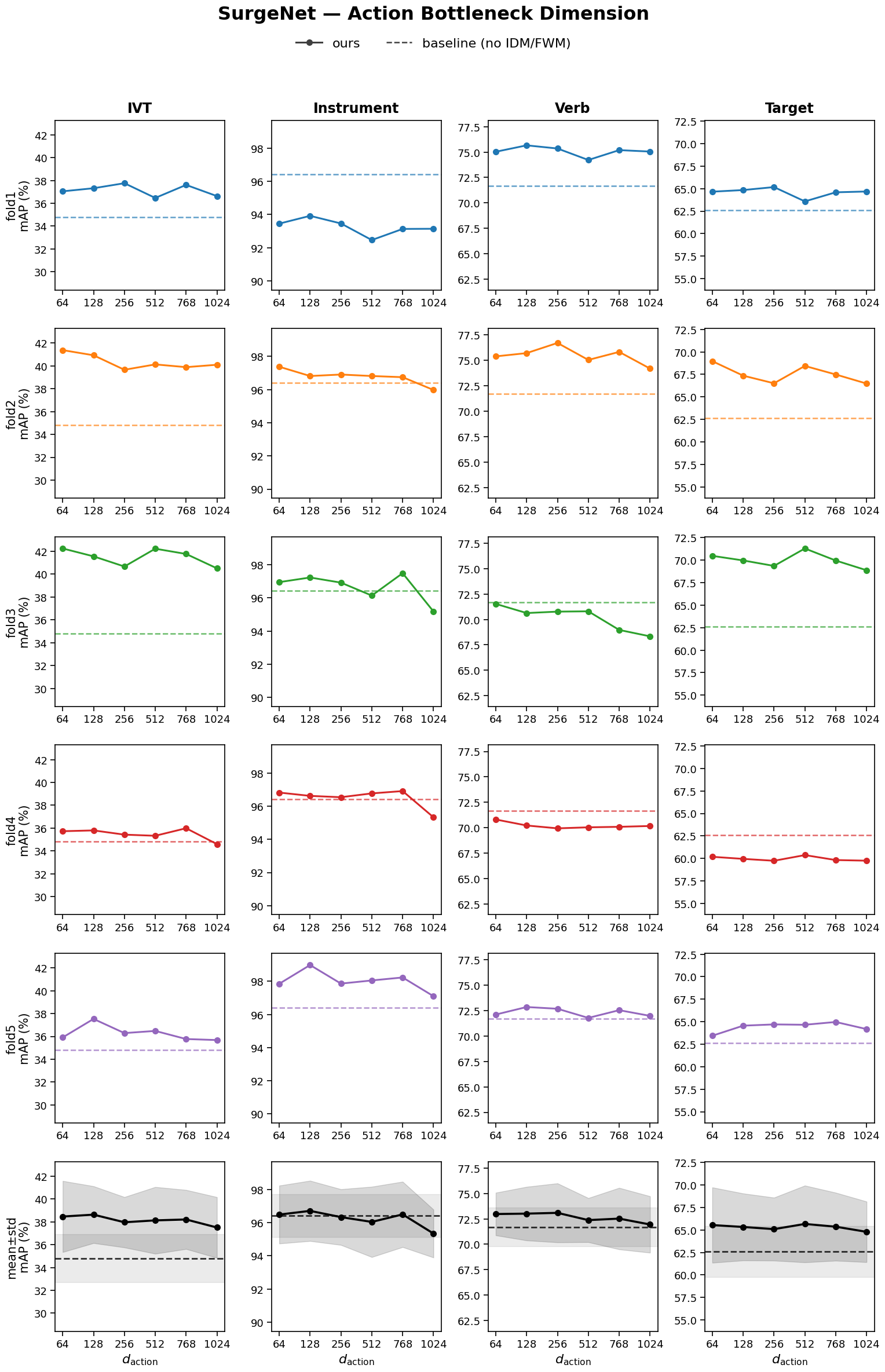}
    \caption{
    \textbf{Effect of latent-action dimension on ProstaTD: SurgeNet-L.}
    Columns report test $\mathrm{mAP}_{IVT}$, $\mathrm{mAP}_{I}$,
    $\mathrm{mAP}_{V}$, and $\mathrm{mAP}_{T}$ across
    $d_a\in\{64,128,256,512,768,1024\}$.
    The first five rows show individual cross-validation folds;
    the bottom row shows their mean with shaded standard deviation.
    Dashed lines indicate the corresponding no-IDM+FDM baseline.
    }
    \label{fig:prostatd_daction_surgenet}
\end{figure}

  \paragraph{Effect of latent-action dimension on ProstaTD.}
Figures~\ref{fig:prostatd_daction_vitl},
\ref{fig:prostatd_daction_dinov2}, and
\ref{fig:prostatd_daction_surgenet} report recognition performance
across $d_a\in\{64,128,256,512,768,1024\}$ for ViT-L,
DINOv2-L, and SurgeNet-L, respectively. Reported per instrument (I), verb (V),
  target (T), and triplet (IVT). Coloured curves are the five cross-validation
  folds; the bottom row of each encoder block is the 5-fold mean $\pm$ std
  (shaded). Dashed lines mark the corresponding no-IDM+FDM baseline. Across ProstaTD encoders, recognition remains broadly stable across
$d_a$, with most changes falling within the five-fold variation.
This contrasts with the stronger capacity-dependent responses observed
on CholecT50 and indicates that the effect of action capacity is
dataset- and encoder-dependent rather than governed by a universal
monotonic trend.

\subsection{More PCA-RGB examples}
\begin{figure}[t]
    \centering
    \includegraphics[width=\linewidth]{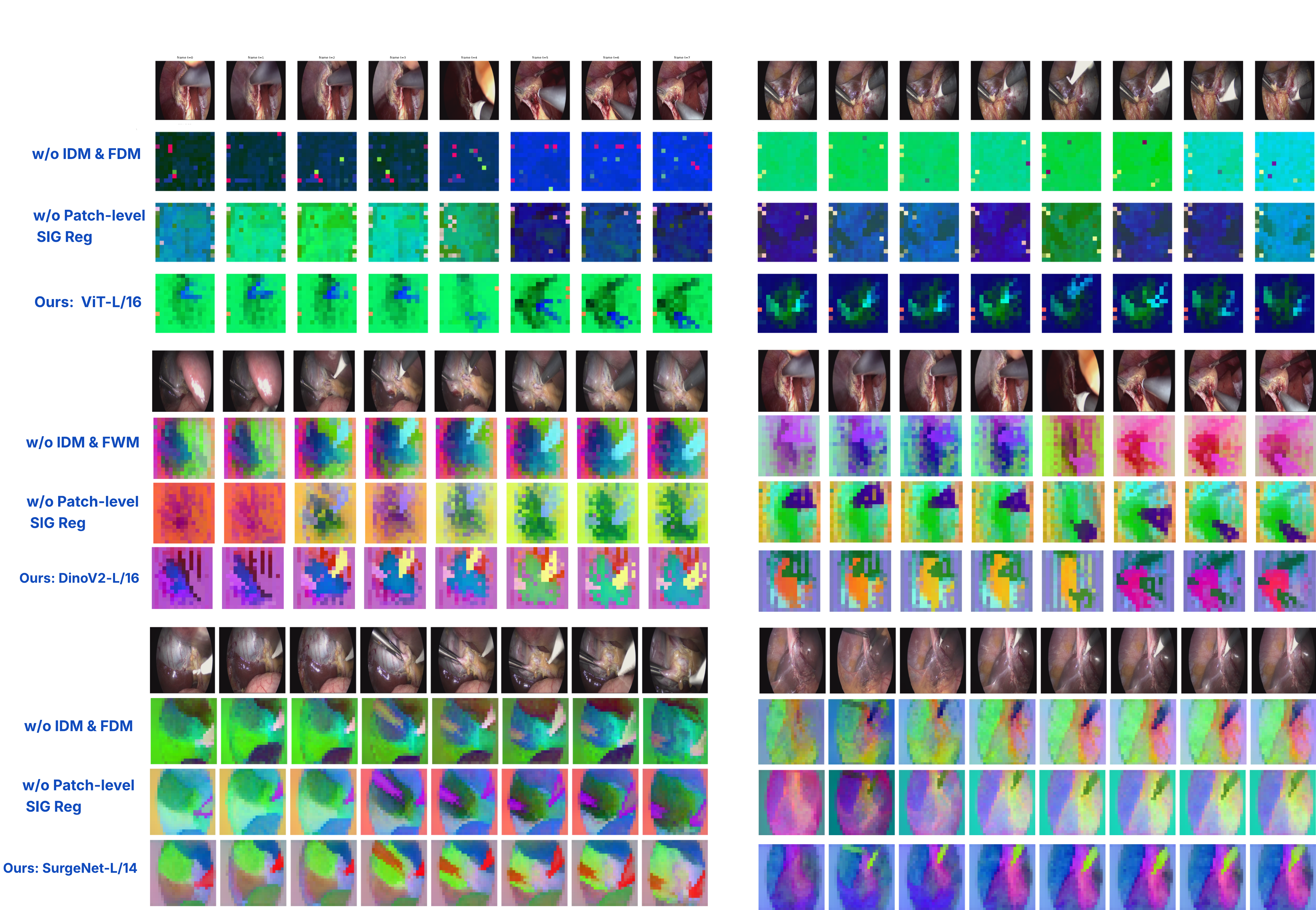}
    \caption{
    PCA-RGB comparison of the full LAViF model ($d_a=128$) ViT-L, DinoV2 and SurgeNet-L, LAViF without
    patch-level SIGReg, and the no-IDM+FDM baseline across the three evaluated
    encoders. The comparison illustrates how latent-action modeling and
    patch-level regularization reshape local patch representations.
    }
    \label{fig:full_ablation_pca}
\end{figure}

\paragraph{PCA-RGB across model components.}
Figure~\ref{fig:full_ablation_pca} compares the full model at $d_a=128$
with the no-patch-SIGReg and no-IDM+FDM variants. Across encoders,
latent-action modeling visibly reorganizes patch-level representations, while
patch-level SIGReg further affects the preservation and separation of local
visual structure.

\begin{figure}[!htbp]
    \centering
    \includegraphics[width=\linewidth]{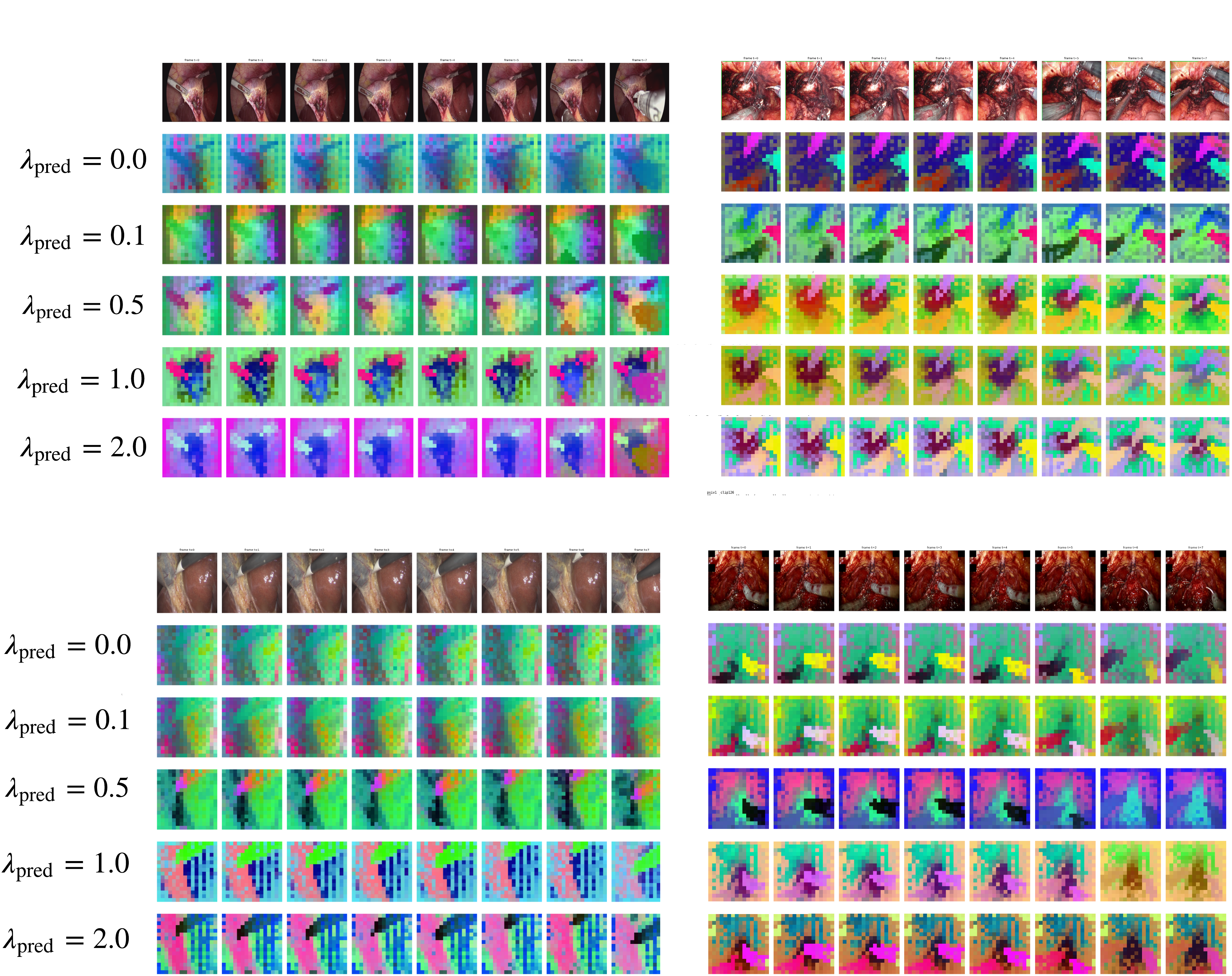}
    \caption{PCA-RGB visualization across different
    $\lambda_{\mathrm{pred}}$ values for DINOv2-L.}
    \label{fig:lambda_pca_dino}
\end{figure}

\begin{figure}[!htbp]
    \centering
    \includegraphics[width=\linewidth]{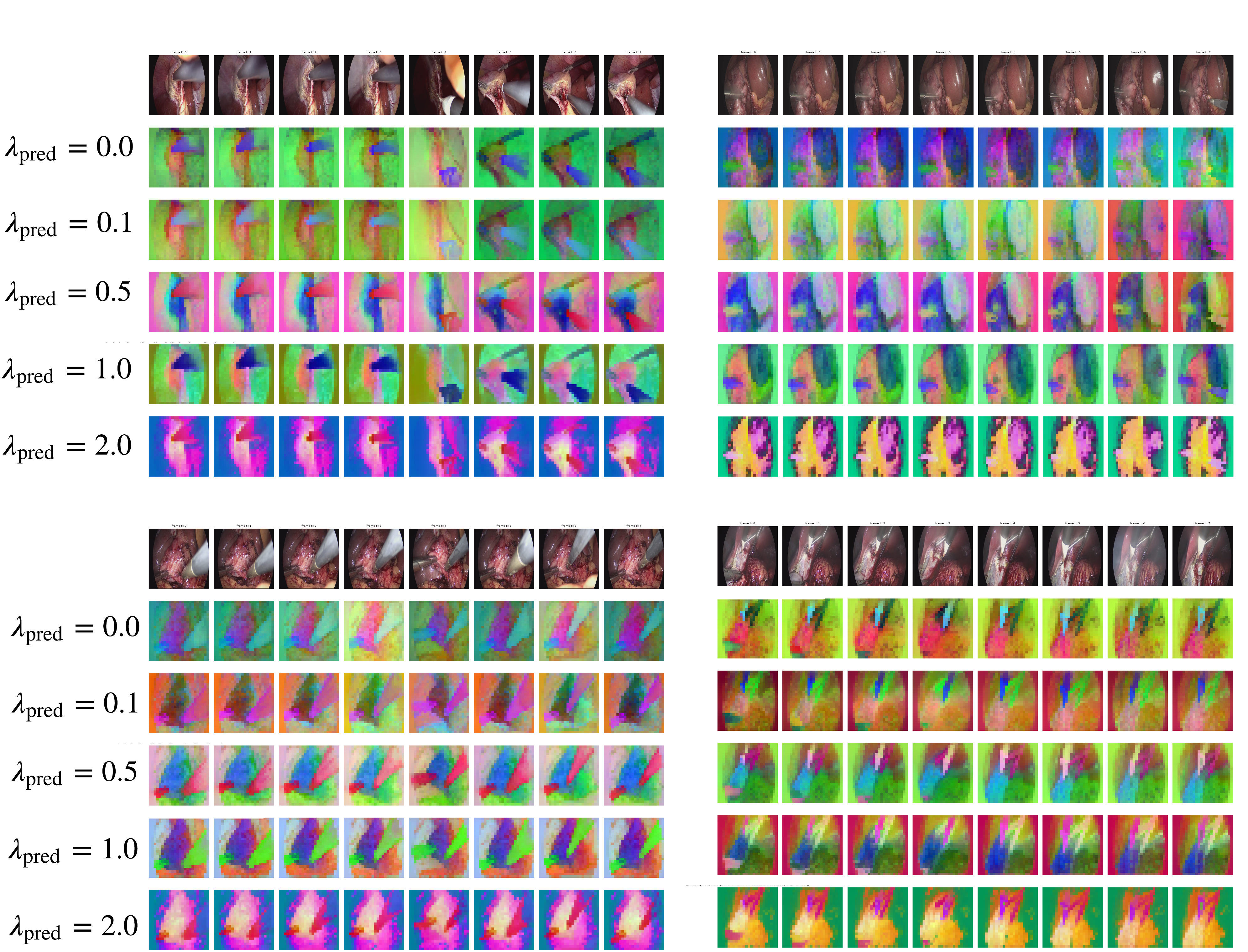}
    \caption{PCA-RGB visualization across different
    $\lambda_{\mathrm{pred}}$ values for SurgeNet-L.}
    \label{fig:lambda_pca_surgnet}
\end{figure}

\begin{figure}[!htbp]
    \centering
    \includegraphics[width=\linewidth]{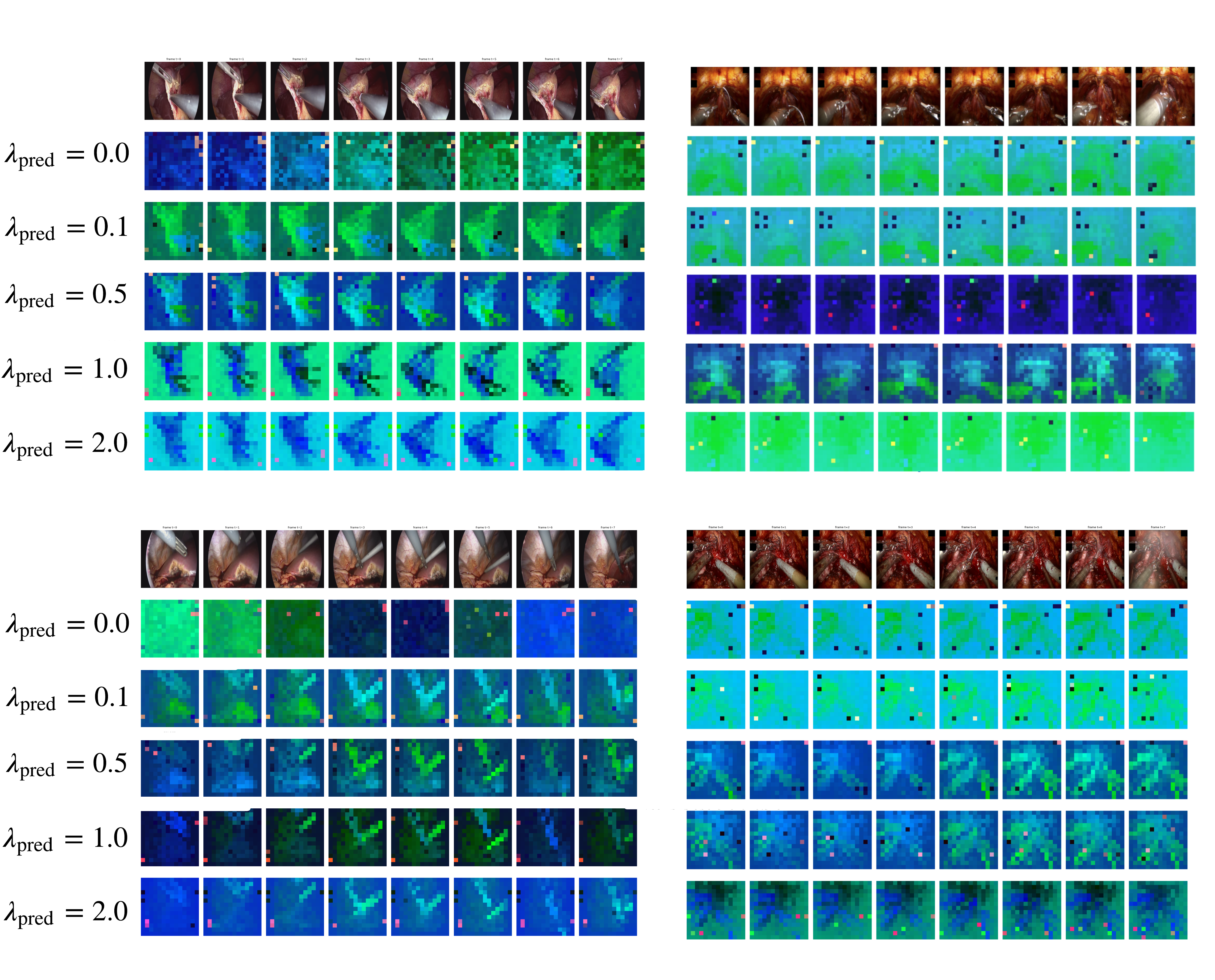}
    \caption{PCA-RGB visualization across different
    $\lambda_{\mathrm{pred}}$ values for ViT-L.}
    \label{fig:lambda_pca_vit}
\end{figure}

\paragraph{PCA-RGB across prediction weights.}
Figures~\ref{fig:lambda_pca_dino}--\ref{fig:lambda_pca_vit} show how
patch-level representations evolve as $\lambda_{\mathrm{pred}}$ varies.
Predictive pressure visibly reorganizes local feature structure for all three
encoders, although the magnitude and spatial pattern of the change are
encoder dependent.

\FloatBarrier

\FloatBarrier

\begin{figure}[t]
    \centering
    \includegraphics[width=\linewidth]{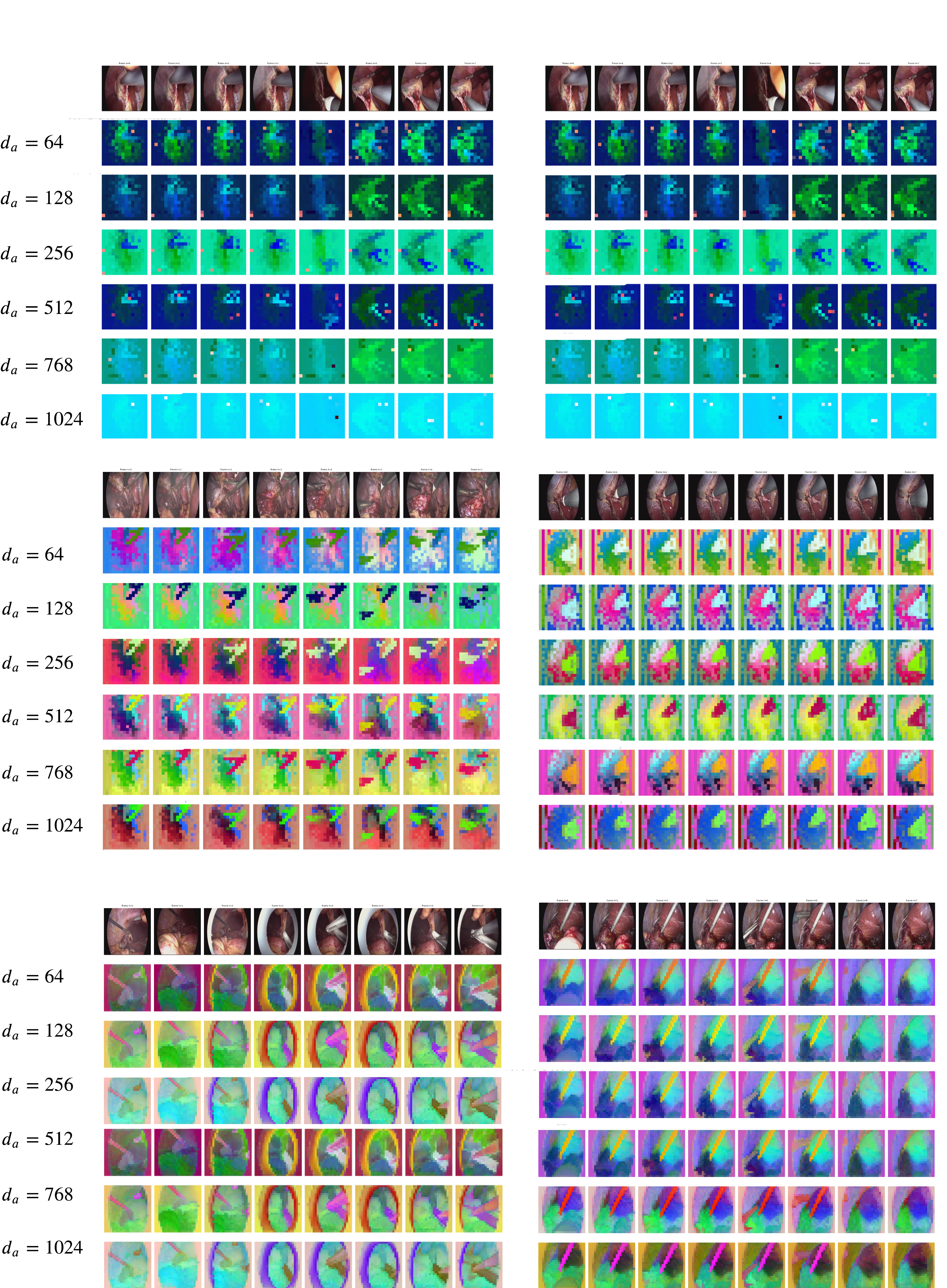}
    \caption{
    PCA-RGB visualization across latent-action dimensions
    $d_a\in\{64,128,256,512,768,1024\}$ for the three evaluated encoders.
    Changing the action bottleneck capacity produces encoder-dependent
    reorganization of patch-level feature structure.
    }
    \label{fig:da_swap_pca}
\end{figure}

\paragraph{PCA-RGB across latent-action capacities.}
Figure~\ref{fig:da_swap_pca} visualizes how patch-level representations change
as the latent-action dimension $d_a$ varies from 64 to 1024 across the three
encoders. Different bottleneck capacities lead to visible reorganization of
local feature structure, with the extent and spatial pattern of these changes
depending on the encoder. This further suggests that $d_a$ controls not only
the capacity of the latent-action channel, but also how visual transition
information is organized in the learned representation.

\subsection{Camera Motions}
To quantify whether the leading directions of the frame-to-frame representation change encode global motion (camera/egomotion) or localized action, we analyze the covariance of the state difference $\delta = s_{t+1}-s_t$. Given an encoder, we collect $\delta$ over $20{,}000$ frame pairs (states are mean-pooled patch tokens, matching training) and take the top eigenvectors ${v_k}$ of $\mathrm{Cov}(\delta)$ — the dominant directions, ordered by variance. For each dominant $v_k$ and each evaluation frame, we back-project it onto the per-patch changes: $a_p = v_k^\top \delta\phi_p$, where $\delta\phi_p = \phi_t[p]-\phi_{t-1}[p]$ is the change of patch $p$ (there are $N$ patches). We then measure how uniform this response is across the frame:
\begin{equation}
U(v_k)=
\frac{\left(\frac{1}{N}\sum_p a_p\right)^2}
     {\frac{1}{N}\sum_p a_p^2}
\in [0,1].
\end{equation}
Intuitively, high uniformity indicates that the dominant
state changes are driven by global camera motion rather than
localized surgical action. We report $U$ averaged over the top-$16$ (and top-$64$) dominant eigenvectors and over $5{,}000$ evaluation frames (per video, then across videos for the multi-fold datasets). Intuitively, a high dominant-direction uniformity means the model's largest-variance state changes are driven by whole-frame camera motion rather than by the surgical action, so the leading directions cannot localize on the instrument/hand, we report our results in Table \ref{tab:uniformity}.
\begin{table}[t]
\centering
\caption{Dominant-direction uniformity of $\mathrm{Cov}(\delta)$ eigenvectors
(lower $=$ more action-localized; higher $=$ camera/global-motion uniform).
Ours (IDM+FDM) vs.\ baseline (no IDM+FDM). ProstaTD is 5-fold (mean$\pm$std);
CholecT50 is single-split. Video encoders are Gemma baselines only.}
\label{tab:uniformity}
\small
\setlength{\tabcolsep}{5pt}
\begin{tabular}{llcccc}
\toprule
& & \multicolumn{2}{c}{top-16 (dominant)} & \multicolumn{2}{c}{top-64} \\
\cmidrule(lr){3-4}\cmidrule(lr){5-6}
Dataset & Encoder & Baseline & Ours & Baseline & Ours \\
\midrule
\multirow{6}{*}{CholecT50}
 & ViT-L        & 0.242 & \textbf{0.173} & 0.194 & \textbf{0.153} \\
 & DINOv2       & 0.381 & \textbf{0.198} & 0.316 & \textbf{0.102} \\
 & SurgeNet     & 0.156 & \textbf{0.062} & 0.159 & \textbf{0.059} \\
 & TimeSformer  & 0.273 & --            & 0.235 & --            \\
 & V-JEPA 2     & 0.162 & --            & 0.091 & --            \\
 & V-JEPA 2.1   & 0.125 & --            & 0.089 & --            \\
\midrule
\multirow{6}{*}{ProstaTD}
 & ViT-L        & $0.288{\scriptstyle\pm0.055}$ & $\mathbf{0.124{\scriptstyle\pm0.019}}$ & $0.241{\scriptstyle\pm0.026}$ & $\mathbf{0.125{\scriptstyle\pm0.012}}$ \\
 & DINOv2       & $0.316{\scriptstyle\pm0.015}$ & $\mathbf{0.226{\scriptstyle\pm0.008}}$ & $0.240{\scriptstyle\pm0.017}$ & $\mathbf{0.140{\scriptstyle\pm0.005}}$ \\
 & SurgeNet     & $0.128{\scriptstyle\pm0.021}$ & $\mathbf{0.107{\scriptstyle\pm0.006}}$ & $0.118{\scriptstyle\pm0.017}$ & $\mathbf{0.063{\scriptstyle\pm0.006}}$ \\
 & TimeSformer  & $0.320{\scriptstyle\pm0.010}$ & --                                    & $0.265{\scriptstyle\pm0.006}$ & --                                    \\
 & V-JEPA 2     & $0.118{\scriptstyle\pm0.016}$  & --  & $0.075{\scriptstyle\pm0.014}$  & -- \\
 & V-JEPA 2.1   & $0.088{\scriptstyle\pm0.005}$  & --  & $0.057{\scriptstyle\pm0.004}$  & -- \\
\bottomrule
\end{tabular}
\end{table}

% \begin{table}[t]
% \centering
% \small
% \setlength{\tabcolsep}{4pt}
% \renewcommand{\arraystretch}{1.05}
% \caption{EK100 multi-instance retrieval results with DINOv2.}
% \label{tab:ek100_pred_ablation}
% \begin{tabular}{c|ccc|ccc|cc}
% \toprule
% $\lambda_{\mathrm{pred}}$
% & \multicolumn{3}{c|}{mAP}
% & \multicolumn{3}{c|}{nDCG}
% & Verb & Noun \\
% & V$\rightarrow$T & T$\rightarrow$V & Avg.
% & V$\rightarrow$T & T$\rightarrow$V & Avg.
% & mAP & mAP \\
% \midrule
% 1.0
% & \textbf{60.84} & \textbf{48.56} & \textbf{53.70}
% & \textbf{66.98} & \textbf{63.29} & \textbf{65.13}
% & \textbf{60.76} & 19.19 \\
% 0.0
% & 58.63 & 47.49 & 53.06
% & 65.84 & 62.88 & 64.36
% & 58.31 & \textbf{19.66} \\
% \bottomrule
% \end{tabular}
% \end{table}
%Only sortedness and nonnegativity of the eigenvalues are used, so the
%equivalence holds for any covariance spectrum, independent of how the
%encoder was trained. The identification of $\rho S$ with the prediction
%loss $\Lp$ in the main text is where the linear LAM analysis of Zhang et
%al.\ enters: their Props.~4.1--4.2, applied to the feature process
%$s_{t+1} = s_t + \delta_t$, show that the optimal linear IDM+FDM pair with
%bottleneck width $\da$ attains exactly the eigenvalue tail
%$\sum_{i>\da}\lambda_i$; for our Transformer IDM+FDM this holds at the
%level of the best linear substructure, so the identification is exact in
%the linear class and an approximation for the trained nonlinear modules.
\begin{table}[t]
\centering
\caption{Relative mAP drop (\%) under temporal reversal. CholecT50 (single split) and ProstaTD (mean$\pm$std over 5 folds). Larger drop $=$ greater reliance on temporal direction.}
\label{tab:reversal_drop2}
\resizebox{\linewidth}{!}{%
\begin{tabular}{ll cccc cccc}
\toprule
& & \multicolumn{4}{c}{CholecT50} & \multicolumn{4}{c}{ProstaTD} \\
\cmidrule(lr){3-6}\cmidrule(lr){7-10}
Encoder & Method & $\Delta IVT$ & $\Delta I$ & $\Delta V$ & $\Delta T$ & $\Delta IVT$ & $\Delta I$ & $\Delta V$ & $\Delta T$ \\
\midrule
\multirow{3}{*}{DINOv2}
& Full \ours{}     & 27.9 & 16.6 & 17.8 & 16.7 & $37.1_{\pm2.4}$ & $19.1_{\pm1.4}$ & $30.0_{\pm2.6}$ & $18.3_{\pm2.7}$ \\
& IDM only           & 23.1 & 15.7 & 16.4 & 14.4 & $37.0_{\pm3.2}$ & $18.9_{\pm1.5}$ & $28.4_{\pm2.4}$ & $18.3_{\pm2.4}$ \\
& No IDM+FDM    & 19.6 & 13.1 & 13.8 & 16.1 & $36.1_{\pm4.6}$ & $19.0_{\pm1.7}$ & $27.8_{\pm3.2}$ & $16.7_{\pm3.6}$ \\
\midrule
\multirow{3}{*}{ViT-L}
& Full \ours{}     & 25.3 & 15.7 & 18.5 & 9.7  & $35.5_{\pm2.8}$ & $17.8_{\pm1.0}$ & $28.6_{\pm2.8}$ & $18.1_{\pm2.2}$ \\
& IDM only           & 21.4 & 14.1 & 16.0 & 9.2  & $31.9_{\pm2.6}$ & $17.1_{\pm1.4}$ & $25.8_{\pm1.5}$ & $16.1_{\pm1.7}$ \\
& No IDM+FDM    & 21.8 & 13.9 & 15.1 & 13.0 & $30.4_{\pm3.3}$ & $16.7_{\pm1.5}$ & $25.0_{\pm1.8}$ & $15.1_{\pm0.8}$ \\
\midrule
\multirow{3}{*}{SurgeNet}
& Full \ours{}     & 32.1 & 18.9 & 21.3 & 14.7 & $37.6_{\pm2.8}$ & $19.9_{\pm1.6}$ & $32.7_{\pm3.0}$ & $18.9_{\pm1.7}$ \\
& IDM only           & 28.7 & 18.6 & 21.3 & 14.2 & $35.1_{\pm3.6}$ & $19.2_{\pm1.5}$ & $28.6_{\pm3.1}$ & $18.3_{\pm0.9}$ \\
& No IDM+FDM    & 28.2 & 15.5 & 19.0 & 14.7 & $34.8_{\pm7.0}$ & $18.9_{\pm2.1}$ & $28.1_{\pm5.1}$ & $16.8_{\pm3.7}$ \\
\midrule
V-JEPA2     & Base & 28.5 & 14.0 & 16.5 & 17.2 & $39.2_{\pm3.4}$ & $20.2_{\pm1.9}$ & $28.1_{\pm3.2}$ & $20.8_{\pm2.5}$ \\
V-JEPA2.1   & Base & 30.0 & 17.7 & 17.0 & 14.4 & $36.9_{\pm5.2}$ & $19.7_{\pm1.2}$ & $28.5_{\pm3.2}$ & $18.6_{\pm2.3}$ \\
TimeSformer & Base & 5.1  & 3.8  & 3.4  & 4.2  & $23.7_{\pm4.2}$ & $12.6_{\pm1.2}$ & $18.7_{\pm2.2}$ & $10.5_{\pm4.1}$ \\
\bottomrule
\end{tabular}%
}
\end{table}

\subsection{Efficiency}
We benchmark end-to-end inference on real CholecT50 clips using a single GPU, reporting per-clip latency, throughput (clips/s), and peak memory (Table~\ref{tab:efficiency}). Because our model encodes each frame independently before the lightweight latent-action aggregation, its cost scales \emph{linearly} in clip length: for a clip of $T$ frames with $N$ patch tokens each, self-attention is applied within each frame, giving $O(T\cdot N^2+T^2)$. Spatiotemporal video encoders (e.g., V-JEPA2/2.1) instead attend jointly over all $T\cdot N$ tokens, so their attention cost is $O(T^2 N^2)$---quadratic in $T$. The gap therefore widens as clips get longer. Empirically, at matched backbone scale ($\sim$300M parameters), our per-frame encoding design reaches $50$--$73$ clips/s, $3.5$--$5\times$ the throughput of the spatiotemporal baselines ($\leq 14.7$ clips/s).

\paragraph{Quantization robustness.}
Beyond throughput, deployability also depends on how well a model tolerates
low-bit weights. We apply post-training, weight-only INT4 quantization to every
linear layer of the trained model, including the vision encoder, IDM, aggregator,
and text encoder. Each weight row is scaled by its maximum absolute value and
rounded to the nearest of 16 integer levels (symmetric, per-output-channel
round-to-nearest), while biases, normalization layers, embeddings, and
activations remain in full precision. No calibration data or fine-tuning is
used, and all models are evaluated on CholecT50 with the same protocol as their
FP32 results (Table~\ref{tab:quant_full}). Our models are nearly unaffected:
triplet mAP changes by $-2.7\%$ (ViT-L), $-2.8\%$ (DINOv2), and $+0.4\%$
(SurgeNet), and the instrument, verb, and target mAPs change by less than $1\%$.
The same encoders trained without IDM+FDM degrade more ($-4.2\%$, $-4.2\%$, and
$-1.3\%$), so training with the latent-action objectives yields representations
that are more robust to weight quantization for every backbone. The
spatiotemporal video baselines are the least robust: triplet mAP drops by
$7.7\%$ (V-JEPA2) and $5.6\%$ (V-JEPA2.1), and verb and target mAP by up to
$4.8\%$. Storing the quantized linear weights in INT4 would shrink them
$8\times$ relative to FP32, further reducing the deployment footprint.

\begin{table}[t]
\centering
\caption{Relative mAP change (\%) under post-training weight-only INT4
quantization on CholecT50, vs.\ FP32. Our models change by at most $2.8\%$ on
triplet mAP, less than the same encoders trained without IDM+FDM and far less
than the spatiotemporal video baselines.}
\label{tab:quant_full}
\setlength{\tabcolsep}{5pt}
\begin{tabular}{llcccc}
\toprule
Encoder & Variant & $\Delta I$ & $\Delta V$ & $\Delta T$ & $\Delta IVT$ \\
\midrule
\multirow{2}{*}{ViT-L}
 & Ours          & $-0.9$ & $-0.8$ & $0.0$  & $-2.7$ \\
 & No IDM+FDM    & $-0.1$ & $-0.5$ & $-2.9$ & $-4.2$ \\
\midrule
\multirow{2}{*}{DINOv2-L}
 & Ours          & $-0.3$ & $-0.2$ & $-0.1$ & $-2.8$ \\
 & No IDM+FDM    & $-0.5$ & $-0.9$ & $-0.9$ & $-4.2$ \\
\midrule
\multirow{2}{*}{SurgeNet-L}
 & Ours          & $-0.1$ & $-0.6$ & $+0.2$ & $+0.4$ \\
 & No IDM+FDM    & $-0.2$ & $-1.0$ & $-0.5$ & $-1.3$ \\
\midrule
V-JEPA2   & Base & $+0.2$ & $-2.4$ & $-3.9$ & $-7.7$ \\
V-JEPA2.1 & Base & $-2.4$ & $-4.8$ & $-4.7$ & $-5.6$ \\
\bottomrule
\end{tabular}
\end{table}

\clearpage
% \bibliographystyle{plain}

% \bibliography{main}
\putbib[main]
\end{bibunit}

\end{document}